\documentclass{article} 
\usepackage{technical_report,times,booktabs,enumitem}

\usepackage{amsmath,amsfonts,bm}

\def\eqref#1{equation~\ref{#1}}

\def\1{\bm{1}}

\DeclareMathAlphabet{\mathsfit}{\encodingdefault}{\sfdefault}{m}{sl}
\SetMathAlphabet{\mathsfit}{bold}{\encodingdefault}{\sfdefault}{bx}{n}

\usepackage{hyperref}
\usepackage{url}
\usepackage{tabularx}
\usepackage{booktabs}
\usepackage{makecell}
\usepackage{subfigure}
\usepackage{graphicx} 
\usepackage{float}
\usepackage{xcolor}
\usepackage{booktabs}
\usepackage{makecell}
\usepackage{array}
\usepackage{multirow}

\newcommand{\eg}{\textit{e.g.}}
\usepackage{caption}
\usepackage{listings}
\usepackage[flushleft]{threeparttable}
\newcolumntype{C}{>{\centering\arraybackslash}X}

\lstdefinestyle{promptstyle}{
  basicstyle=\ttfamily\small,
  frame=single,
  framesep=8pt,
  framerule=0.4pt,
  rulecolor=\color{black!40},
  breaklines=true,
  breakatwhitespace=true,
  showstringspaces=false,
  columns=fullflexible,
  keepspaces=true,
  postbreak=,
}

\title{Logics-Parsing-Omni Technical Report}

\author{
Logics Team  \\
\\
Alibaba Group\\
\\
(For the complete list of authors, please refer to the Contribution section)
}

\iclrfinalcopy 
\usepackage{hyperref}
\usepackage{xcolor}
\usepackage{tikz}
\usepackage{fontawesome5}
\usepackage{lmodern}

\definecolor{hfOrange}{HTML}{FF9D00}
\definecolor{ghBlack}{HTML}{24292F}
\definecolor{arxivRed}{HTML}{B31B1B}

\newcommand{\resourceBar}{%
  \begin{center}%
  \vspace{-30pt}%
  {\scriptsize \faGithub\
  \href{https://github.com/alibaba/Logics-Parsing/tree/main/Logics-Parsing-Omni}%
       {\texttt{github.com/alibaba/Logics-Parsing/tree/main/Logics-Parsing-Omni}}}%
  \\[1pt]
  {\scriptsize \textcolor{hfOrange}{\textbf{Model: }}\
  \href{https://huggingface.co/Logics-MLLM/Logics-Parsing-Omni}%
       {\texttt{huggingface.co/Logics-MLLM/Logics-Parsing-Omni}}}%
  \\[1pt]
  {\scriptsize \textcolor{hfOrange}{\textbf{OmniParsingBench: }}\
  \href{https://huggingface.co/datasets/Logics-MLLM/OmniParsingBench}%
       {\texttt{huggingface.co/datasets/Logics-MLLM/OmniParsingBench}}}%
  \vspace{0pt}%
  \end{center}%
}

\begin{document}
\vspace*{-20pt}
\maketitle
\resourceBar


\begin{abstract} 
Addressing the challenges of fragmented task definitions and the heterogeneity of unstructured data in multimodal parsing, this paper proposes the Omni Parsing framework. This framework establishes a Unified Taxonomy covering documents, images, and audio-visual streams, introducing a progressive parsing paradigm that bridges perception and cognition. Specifically, the framework integrates three hierarchical levels: 1) Holistic Detection, which achieves precise spatial-temporal grounding of objects or events to establish a geometric baseline for perception; 2) Fine-grained Recognition, which performs symbolization (e.g., OCR/ASR) and attribute extraction on localized objects to complete structured entity parsing; and 3) Multi-level Interpreting, which constructs a reasoning chain from local semantics to global logic. A pivotal advantage of this framework is its evidence anchoring mechanism, which enforces a strict alignment between high-level semantic descriptions and low-level facts. This enables ``evidence-based'' logical induction, transforming unstructured signals into standardized knowledge that is locatable, enumerable, and traceable. Building on this foundation, we constructed a standardized dataset and released the Logics-Parsing-Omni model, which successfully converts complex audio-visual signals into machine-readable structured knowledge. Experiments demonstrate that fine-grained perception and high-level cognition are synergistic, effectively enhancing model reliability. Furthermore, to quantitatively evaluate these capabilities, we introduce OmniParsingBench.
\end{abstract}

\begin{figure}[htbp]
    \centering
    \includegraphics[width=1.0\textwidth]{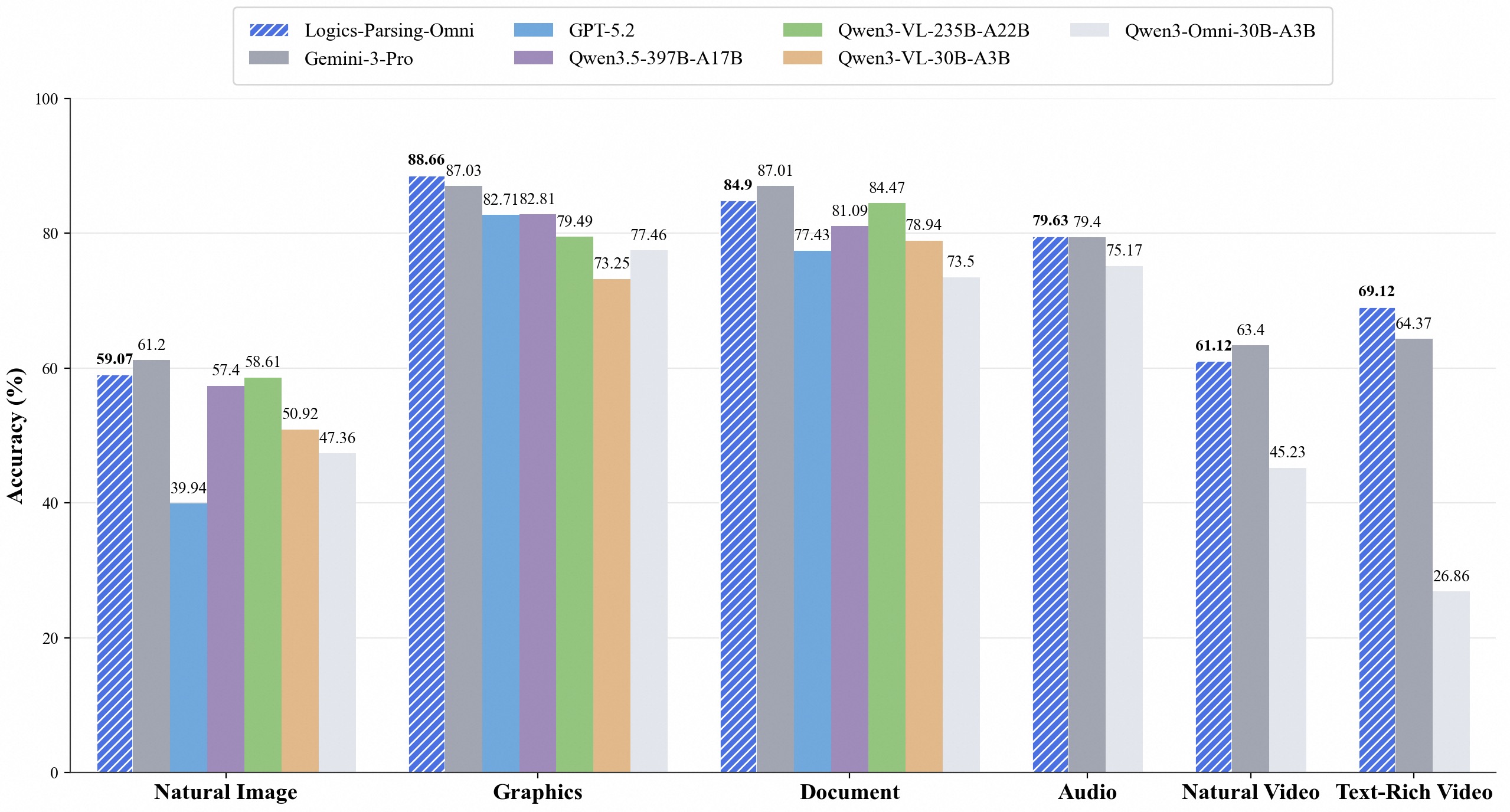}
    \caption{OmniParsingBench performance of Logics-Parsing-Omni.}
    \label{fig:bar_doc_omni}
\end{figure}

\begin{figure}[htbp]
    \centering
    \includegraphics[width=1.0\textwidth]{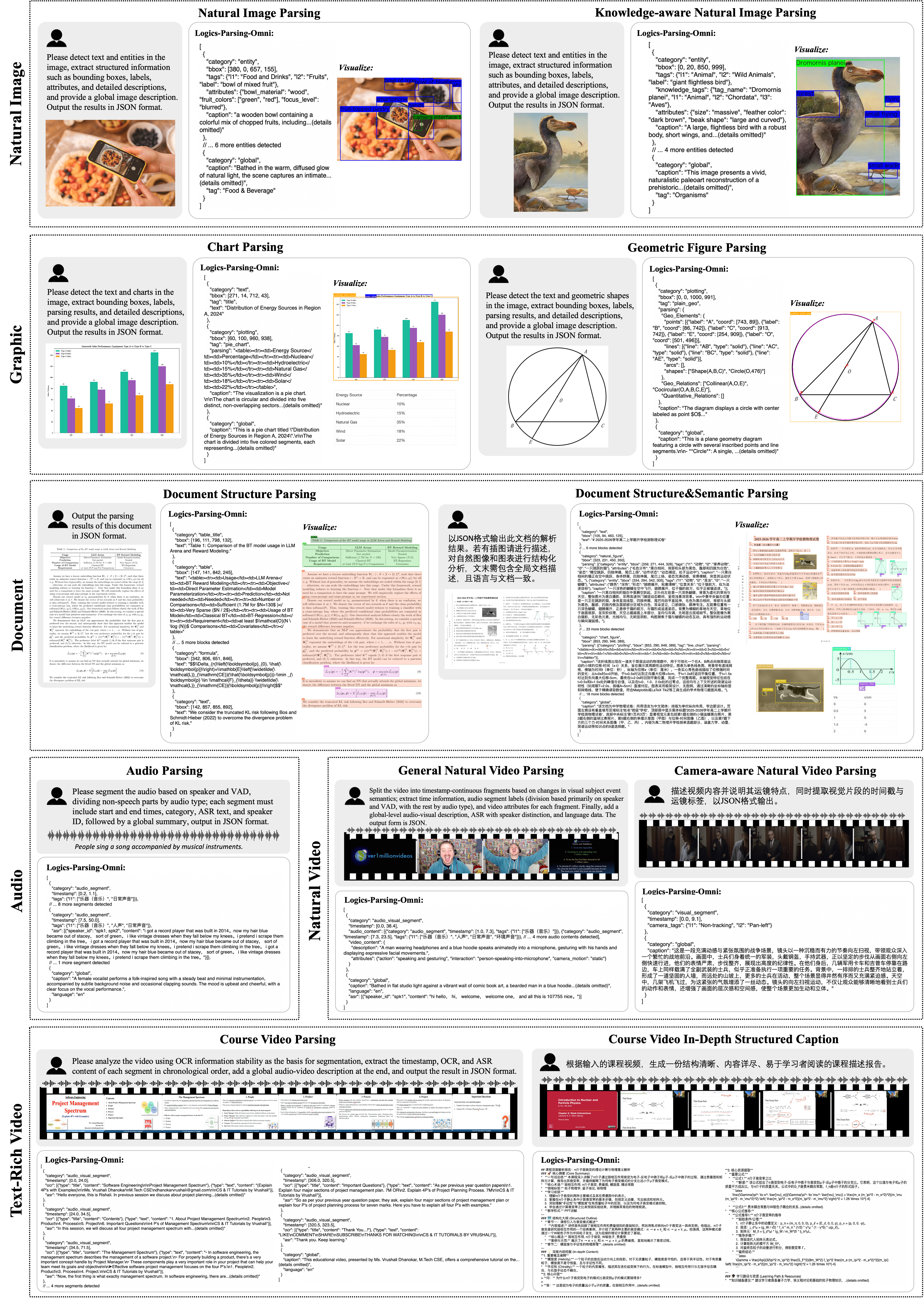}
    \caption{Showcase of the multifaceted capabilities of Logics-Parsing-Omni.}
    \label{fig:showcase_all}
\end{figure}

\section{Introduction} 
In recent years, Multimodal Large Language Models (MLLMs) have demonstrated remarkable general-purpose capabilities \cite{liu2023llava,LLaVA-OneVision-1.5,Qwen3-VL,yang2025survey}. However, applying them to knowledge-intensive domains remains challenging, particularly in visual-rich documents and long-form educational videos. These media feature high-density text, complex layouts, and diverse non-text visual content (e.g., illustrations, charts, and slides). For practical deployment, neither low-level extraction (e.g., OCR or layout analysis) nor free-form high-level semantic descriptions alone are sufficient. An effective system requires the integration of fine-grained content parsing with deep semantic understanding to produce a unified representation that is locatable, indexable, and summarizable—thereby enabling robust downstream tasks such as Retrieval-Augmented Generation (RAG), Question Answering (QA), and intelligent tutoring.

Current approaches face structural dilemmas across modalities. In the document domain, traditional pipelines \cite{wei2024general,poznanski2025olmocr,cui2025paddleocr} excel at layout analysis but often reduce complex figures and charts to mere bounding boxes, stripping them of semantic value. This leads to the loss of critical information such as trends, comparisons, causal relations, and chart insights, rendering them unusable for deep retrieval. While LLM-based approaches \cite{wang2024qwen2,bai2025qwen2} generate fluent descriptions, they often lack layout fidelity and fine-grained grounding, limiting their utility in automated document conversion and indexing pipelines. 
Similarly, in image understanding, generalist models often fail to capture the dense, attribute-rich information (e.g., specific chart data or spatial relations) required for complex reasoning and tend to hallucinate or gloss over these details in favor of generic descriptions. The audio-visual domain faces even greater hurdles: relying solely on linear ASR transcripts overlooks non-speech acoustic events (e.g., background noises, emotional cues)  and visual contexts (e.g., slides), while generic video captions \cite{xu2025qwen2,Qwen3-Omni,ai2025ming,team2025longcat}  lack the structural granularity to organize long-form content. Consequently, there is a critical need for a system that unifies reliable, fine-grained parsing with application-oriented semantic understanding.

To this end, we propose the Omni Parsing framework, establishing a unified parsing taxonomy across documents, images, and audio-visual streams. Unlike traditional paradigms treating perception and cognition separately, Omni Parsing introduces a progressive paradigm to bridge pixel-based perception and logic-based cognition. We define ``Omni Parsing'' as transforming unstructured signals into standardized knowledge that is Locatable, Enumerable, and Traceable. This ensures that the understanding of visual elements—whether document diagrams or video events—is not merely descriptive but forms an indispensable part of a structured, fact-based reasoning chain.

Building on this framework, we release Logics-Parsing-Omni, an advanced MLLM designed to instantiate this progressive parsing paradigm across all modalities. The model is optimized for three synergistic tasks: Holistic Detection for spatiotemporal localization; Fine-grained Recognition for symbol and attribute extraction; and Semantic Interpretation for logical reasoning.  Crucially, the model's performance is driven by a data-centric strategy. We significantly enriched knowledge-intensive image samples to enhance entity-rich reasoning. Furthermore, for the video domain, we specifically optimized annotations for fine-grained shot analysis and long educational content, ensuring accurate temporal localization and narrative understanding. This approach effectively anchors generation in facts, resulting in semantically rich and verifiable outputs. To provide a quantitative overview of omni-modal parsing, we built an internal OmniParsingBench covering document, image, and audio-video content. Figure~\ref{fig:bar_doc_omni} compares Logics-Parsing-Omni with representative baseline models on OmniParsingBench. The results show consistent improvements across all modalities, demonstrating that the model maintains a strong balance between structural fidelity and semantic interpretation. Beyond quantitative results, Figure~\ref{fig:showcase_all} provides a compact qualitative showcase of our unified parsing across modalities.

The main contributions of our work can be summarized as follows: 
\begin{itemize}[topsep=0pt, partopsep=0pt, itemsep=2pt, parsep=0pt, leftmargin=*]
\item Unified Framework: We propose the Omni Parsing framework, defining a three-level progressive paradigm—integrating Holistic Detection, Fine-grained Recognition, and Semantic Interpretation—to unify perception and cognition across documents, images, and audio-visual streams.
\item Data \& Methodology: We constructed a standardized, high-quality dataset tailored for omni-modal parsing. By integrating knowledge-intensive image data and optimizing video annotations for structural analysis, we enable the model to master both complex static entities and dynamic narrative logic.
\item Model \& Benchmark: We release Logics-Parsing-Omni, a model capable of converting complex multimodal signals into machine-readable knowledge, and open-source OmniParsingBench to facilitate the quantitative evaluation of fine-grained parsing capabilities, providing robust infrastructure for the community.
\end{itemize}

\section{Methodology} 
In this section, we present the methodology of Logics-Parsing-Omni. Leveraging data structured by our three-level progressive parsing framework, we employ a two-stage training paradigm designed to seamlessly connect fine-grained perception with high-level cognition. We first introduce the overall framework (Section~2.1), then describe the construction of training data (Section~2.2), and finally detail the training procedure (Section~2.3).

\subsection{Overview}
\begin{figure}[htbp]
    \centering
    \includegraphics[width=1.0\textwidth]{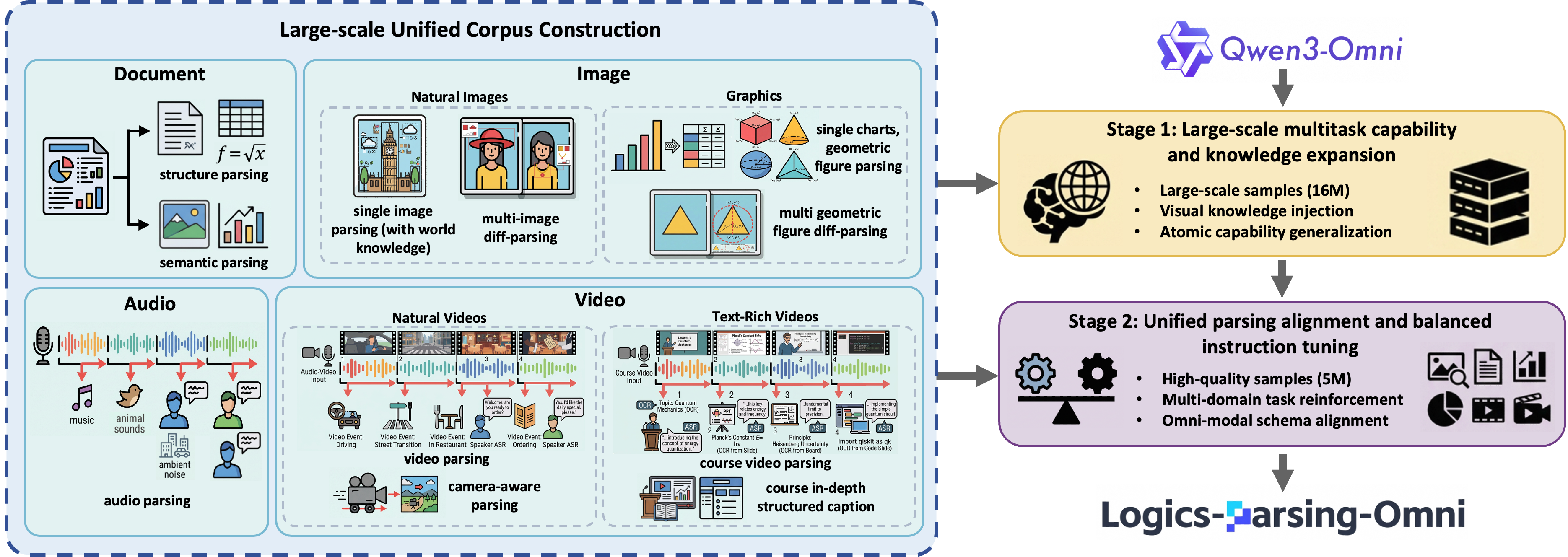}
    \caption{The construction of unified multi-modal parsing corpus and training pipeline of our proposed Logics-Parsing-Omni.}
    \label{fig:framework}
\end{figure}

As illustrated in Figure~\ref{fig:framework}, the foundation of our approach is the construction of a large-scale unified corpus that systematically integrates heterogeneous tasks across four core modalities: Document, Image, Audio, and Video. Rather than treating these modalities in isolation, we unify diverse objectives—ranging from structural and semantic parsing in documents, to single and multi-image difference analysis (across natural scenes and graphics), as well as complex temporal understanding in diverse audio and video environments. By formulating these multifaceted tasks into a cohesive data pipeline, we encourage the model to jointly learn localized structural fidelity and high-level semantic logic.

We adopt a two-stage progressive training strategy. The first stage, Panoramic Cognitive Foundation, prioritizes data scale and coverage to foster broad visual knowledge and atomic capabilities. Through exposure to diverse multi-source tasks, the model masters pure structural parsing and isolated semantic recognition without being overwhelmed by complex reasoning. 
Building upon this, the second stage, Unified Parsing Alignment, refines the model using high-quality, balanced instructions. This progression ensures the model maps heterogeneous omni-modal inputs into standardized JSON formats while preserving fluent natural language generation, achieving a deep integration of perception and cognition.

\subsection{Dataset} 
\begin{figure}[htbp]
    \centering
    \includegraphics[width=1.0\textwidth]{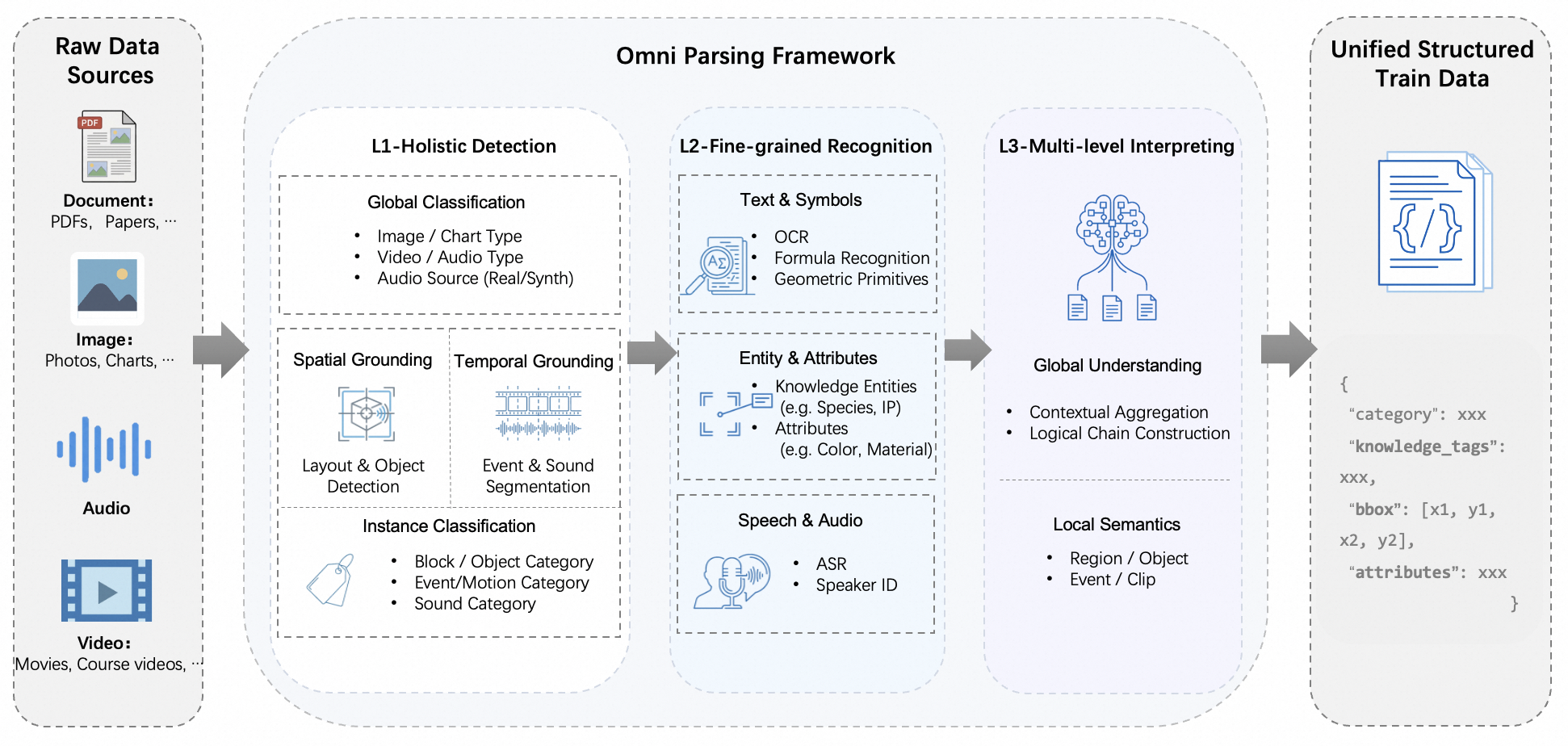}
    \caption{Overview of the Omni Parsing Framework.
The framework transforms multimodal raw data into unified structured training data via three progressive stages:
(1) L1-Holistic Detection: Performs spatio-temporal grounding and coarse classification;
(2) L2-Fine-grained Recognition: Extracts detailed text, symbols, knowledge, attributes, and speech content;
(3) L3-Multi-level Interpreting: Synthesizes local semantics with global logical reasoning.
The final output is a standardized JSON format containing all parsing results from L1 to L3.}
    \label{fig:data_synthesis_pipeline}
\end{figure}

As illustrated in Figure~\ref{fig:data_synthesis_pipeline}, we construct a large-scale, diverse, and high-quality corpus for unified parsing across modalities. To support the progressive training strategy in Section~2.1 --- including both layout-faithful structural parsing and semantic interpretation of visual elements, the corpus covers four domains: Image, Document, Audio and Video: 
\begin{itemize}[topsep=0pt, partopsep=0pt, itemsep=2pt, parsep=0pt, leftmargin=*]
\item \textbf{Image.} This domain encompasses both natural images and information-centric graphics (e.g., charts and geometric figures). It explicitly transforms visual content into standardized, machine-readable formats with precise spatial localization, fine-grained semantic interpretation, and reliable knowledge association.
\item \textbf{Document.} We adopt a unified schema for the layout-faithful extraction of heterogeneous elements, including text, tables, and mathematical formulas. Crucially, it moves beyond traditional OCR by contextually interpreting embedded visual elements (e.g., illustrations) into searchable semantic descriptions, ensuring comprehensive document understanding.
\item \textbf{Audio.} Moving beyond treating audio merely as a source of speech, we decode it as a multi-dimensional sensory channel. This domain provides precise, time-aligned semantic chunks that seamlessly integrate speaker-attributed transcriptions with fine-grained acoustic events and overarching scene descriptions.
\item \textbf{Video.} Extending spatial parsing into the temporal dimension, this domain hierarchically fuses visual dynamics, audio signals, and explicit camera motion. It also encompasses text-rich instructional videos, effectively converting complex, long-form audio-visual streams into temporally aligned, structured multimodal knowledge.
\end{itemize}

The detailed statistics are summarized in Table~\ref{tab:dataset_statistics}. The corpus is constructed as a hybrid mixture across all four modalities: it includes not only the unified structured parsing data generated by our Omni Parsing framework, but also diverse Caption and QA data. This specific data composition directly fuels our two-stage training strategy. It provides the 16M large-scale knowledge coverage necessary to establish atomic capabilities and basic perception in Stage 1, while offering the 5M high-precision instructional support required for deep balanced fine-tuning and complex reasoning in Stage 2. Ultimately, this ensures the model achieves consistent full-modal alignment from basic structure to advanced semantics.

\begin{table}[htbp]
    \centering
    \fontsize{8.5pt}{10.2pt}\selectfont 
    \caption{Detailed statistics of the constructed corpus.}
    \label{tab:dataset_statistics}
    \renewcommand{\arraystretch}{1.15}
    
    \begin{tabularx}{\linewidth}{ll>{\raggedright\arraybackslash}Xr}
        \toprule
        \textbf{Domain} & \textbf{Data Category} & \textbf{Task} & \textbf{\# Samples} \\
        \midrule
        
        \multirow{2}{*}{Document} 
            & \multirow{2}{*}{General Document} 
            & Parsing (Structure)& 342K \\
            & & Parsing (Structure\&Semantic)& 333K \\
        \midrule
        
        \multirow{13}{*}[-0.6em]{Image} 
            & \multirow{4}{*}{Natural Image} 
            & Caption (Single-Image) & 360K \\
            & & Parsing (Single-Image) & 411K \\
            & & Caption (Diff, Multi-Image) & 354K \\
            & & Parsing (Diff, Multi-Image) & 212K \\
            \cmidrule(l){2-4} 
            
            & \multirow{3}{*}{Knowledge-aware Natural Image} 
            & QA & 12.6M \\
            & & Caption & 400K \\
            & & Parsing & 400K \\
            \cmidrule(l){2-4} 
            
            & \multirow{6}{*}{Graphics} 
            & Charts Caption & 132K \\
            & & Charts Parsing & 176K \\
            & & Geometric Caption (Single-Image) & 266K \\
            & & Geometric Parsing (Single-Image) & 174K \\
            & & Geometric Caption (Diff, Multi-Image) & 444K \\
            & & Geometric Parsing (Diff, Multi-Image) & 358K \\
        \midrule

        \multirow{3}{*}[-0.0em]{Audio} 
            & \multirow{3}{*}{General Audio} 
            & QA & 100K \\
            & & Caption & 511K \\
            & & Parsing & 266K \\
        \midrule
        
        \multirow{9}{*}[-0.6em]{Video} 
            & \multirow{3}{*}{General Natural Video} 
            & QA & 30K \\
            & & Caption & 620K \\
            & & Parsing & 269K \\
            \cmidrule(l){2-4}
            
            & \multirow{3}{*}{Camera-aware Natural Video} 
            & QA & 97K \\
            & & Caption & 47K \\
            & & Parsing & 47K \\
            \cmidrule(l){2-4}
            
            & \multirow{3}{*}{Text-Rich Video} 
            & Caption (General) & 90K \\
            & & Caption (In-Depth) & 130K \\
            & & Parsing & 79K \\
        \bottomrule
    \end{tabularx}
\end{table}

\subsubsection{Image}
\paragraph{Natural Images.} \label{sec:nature_image_dataset} 
To address the limitations of traditional captioning, such as subjective ambiguity,  absence of spatial localization, coarse semantic granularity, and unstructured expression, we propose a structured parsing approach that deconstructs images into a standardized JSON format.

\begin{itemize}[topsep=0pt, partopsep=0pt, itemsep=2pt, parsep=0pt, leftmargin=*]
\item \textbf{Single-Image.} This module transforms visual content into logically coherent, machine-friendly structured data. By explicitly anchoring semantics to spatial regions while preserving readability, it significantly enhances utility in fine-grained retrieval, cross-modal alignment, and accessibility services.

\begin{itemize}[topsep=0pt, partopsep=0pt, itemsep=2pt, parsep=0pt, leftmargin=*]
\item \textbf{Parsing Format:} The output follows a unified schema comprising three structured components: Entity Objects (encapsulating attributes, normalized coordinates, and objective descriptions), Text Blocks (preserving layout, functional categories, and multilingual transcriptions), and Global Descriptions (integrating scene classification with a cohesive spatial narrative). This standardized format ensures all visual elements are locatable and interpretable, facilitating flexible integration into downstream tasks.

\item \textbf{Knowledge-aware Entity Parsing:} To bridge the gap between generic descriptions and specific identities, we introduce a strict Knowledge Enhancement Mechanism. Entities are linked to authoritative identifiers (e.g., landmarks, brands, species) solely when visual evidence is unambiguous. By integrating these verified knowledge tags into the narrative, we transform unstructured signals into precise, actionable knowledge, essential for applications like semantic search.

\item \textbf{Data Preparation:} The dataset is constructed via a collaborative pipeline combining expert model synthesis (Detection, OCR) with rule-based standardization. Knowledge tags are injected exclusively for validated entities, while global descriptions are refined by an LLM that synthesizes multi-perspective drafts into a coherent narrative. The final corpus undergoes automated validation to ensure high quality.
\end{itemize}

\item \textbf{Multi-Image.}
To enhance the model’s capability in recognizing fine-grained differences in natural images, we constructed a high-quality Image Difference Dataset. We formed image pairs from real-world collections, applying rigorous VLM-based filtering to ensure consistency and aesthetic quality. The annotation process employs a fully automated closed-loop pipeline: a VLM first analyzes semantic discrepancies to generate descriptive captions, which then prompt a Grounding VLM to precisely localize differing regions. This ensures strict alignment between linguistic descriptions and geometric grounding. By leveraging real-world imagery rather than synthetic artifacts, our dataset provides high-fidelity supervision for fine-grained visual comparison tasks, establishing a reliable benchmark for training and evaluating models.
\end{itemize}

\paragraph{Graphics.} \label{para:graphics} To enhance the model's capability in accurately interpreting and reasoning over information-dense visual elements, we curated a specialized structural repository focusing on charts and geometric figures. 

\begin{itemize}[topsep=0pt, partopsep=0pt, itemsep=2pt, parsep=0pt, leftmargin=*]
\item \textbf{Charts.} We integrated diverse chart samples from multiple public datasets~\cite{li2024llava, hu2024mplug, pritamdeka_mermaid2024, sroecker_mermaidflowchart2024, kingsoft_qzhou2024, tbt0205_flowchart42024} (\eg, line/bar/pie charts, flowcharts, infographics). To align with the Omni Parsing framework, we annotated these data following our three-level progressive paradigm:
In the \textit{L1 \& L2 stages (Detection \& Recognition)}, we decouple plotting regions from auxiliary text regions. Texts are extracted and classified by functional tags (\eg, titles), while plotting regions undergo fine-grained categorization to ensure all visual entities are locatable.
In the \textit{L3 stage (Semantic Interpreting)}, we utilize Qwen3-VL-235B-A22B-Instruct to reverse-engineer visual elements into precise structured formats. Specifically, statistical charts are parsed into HTML tables (\texttt{<table>...</table>}), while flowcharts and relational diagrams are transcribed into Mermaid code. This ensures high-precision extraction of data series and aligns perfectly with table parsing in the Document domain. Based on this, we generate \textit{Local Captions} detailing chart types, axes, legends, and specific data points, followed by a \textit{Global Caption} that explicitly interprets overarching data trends, distributions, and key annotations.

\item \textbf{Geometric Figures.}
To tackle the complex spatial topology and mathematical constraints of geometric figures, we construct two subsets covering single-image interpretation and multi-image difference parsing:

\begin{itemize}[topsep=0pt, partopsep=0pt, itemsep=2pt, parsep=0pt, leftmargin=*]
\item \textbf{Single-Image.} Sourced from K--12 textbooks and public datasets~\cite{li2024llava, peng2024multimath}, we decouple plotting regions from textual contexts (\eg, mathematical problem stems). We synthesize structured data encapsulating three dimensions: \textit{Geometric Elements} (precise localization of points/lines/polygons), \textit{Topological Relations} (\eg, parallelism), and \textit{Quantitative Relations} (\eg, lengths/radii). For plane geometry, the overarching logic is synthesized via Qwen3-VL-235B-A22B-Thinking, with coordinates and radii anchored by a specialized recognition model. Due to severe viewpoint complexities, 3D solid geometry is parsed using Gemini-3-Pro. Finally, we generate \textit{Local Captions} detailing primitives and metric annotations, and \textit{Global Captions} fusing visual logic with the textual context.

\item \textbf{Multi-Image.} We construct minimal-edit image pairs differing by only atomic operations (\eg, drawing an angle bisector). To enable fine-grained reasoning, annotations are structured into two levels: the \textit{Plotting level} specifies differential visual elements (added primitives, coordinates) and constraints (geometric/quantitative), while the \textit{Global level} provides natural-language editing instructions. The English subset is synthesized via an enhanced MathCanvas pipeline, and the Chinese subset is mined from K--12 textbooks using Qwen3-VL. All candidate pairs are rigorously filtered to ensure semantically meaningful and curriculum-aligned transformations.
\end{itemize}
\end{itemize}

\subsubsection{Document} 
For the document parsing task, we choose to leverage the supervised fine-tuning (SFT) dataset originally introduced in our prior \textit{Logics-Parsing}~\cite{chen2025logics} framework. Empirical evaluations have demonstrated that this dataset effectively equips the \textit{Logics-Parsing} model with robust document parsing capabilities, attributed to its comprehensive coverage of heterogeneous document types as well as balanced distribution of English and Chinese content. Concretely, the dataset is constructed from two complementary sources. The first comprises data aggregated from several well-established public datasets, whose original annotations are normalized into a unified JSON annotation schema, including olmOCR-mix-0225~\cite{poznanski2025olmocr} for page-level parsing, FinTabNet~\cite{DBLP:journals/corr/abs-2005-00589}, TNCR~\cite{DBLP:journals/corr/abs-2106-15322}, and PubTabNet~\cite{DBLP:journals/corr/abs-1911-10683} for table recognition sub-task, as well as ChEBI-20-MM dataset~\cite{liu2025quantitativeanalysisknowledgelearningpreferences} for chemical structure recognition. The second source originates from a large-scale in-house dataset processed through a rigorous two-stage annotation pipeline which involves automated pre-annotation and expert refinement (both model-assisted and human-verified). In total, more than 300K high-quality page-level document images are collected for document parsing task.



Beyond traditional OCR and layout analysis, we explicitly integrate fine-grained structured parsing for embedded non-text elements (e.g., illustrations, charts). This unified modeling captures text-vision logical connections, unlocking the semantic value of figures for deep retrieval.


\subsubsection{Audio} 
\label{sec:method_dataset_audio}


To decode the multifaceted signals within the audio modality, we construct a time-aligned parsing process that integrates non-dialogue acoustic events with speaker identification to support high-fidelity cross-modal inference. 

\begin{itemize}[topsep=0pt, partopsep=0pt, itemsep=2pt, parsep=0pt, leftmargin=*]
\item \textbf{Speaker-Attributed Transcription.} Utilizing the CAM++ \cite{wang2023camfastefficientnetwork} model, we perform speaker diarization to assign unique speaker IDs to distinct audio segments. These segments are transcribed using FireRedASR \cite{xu2025fireredasropensourceindustrialgrademandarin} and Whisper \cite{radford2022whisper}, establishing a synchronized "SpeakerID-ASR-Timestamp" triplet. This ensures that every utterance is precisely anchored to a specific identity and time interval.
\item \textbf{Acoustic Scene Modeling.} To capture non-linguistic cues, we employ the DASM \cite{cai2025detectsoundopenvocabularysound} framework for dense acoustic event detection. DASM identifies over 500 granular acoustic classes, which we distill into six macro-categories: Human Voice, Instrumental Music, Environmental Noise, Traffic, Animals, and Daily Life Sounds. We optimize classification thresholds to generate high-precision, timestamped acoustic labels.
\item \textbf{Unified Audio Semantic Chunks.} We synthesize these streams by using speaker segments as the primary temporal units, overlaying DASM-detected acoustic events while merging overlapping intervals and removing noise. The resulting Audio Semantic Chunks contain start/end boundaries, Speaker IDs, ASR text, and acoustic event tags, complemented by holistic scene descriptions from MidasHengLM \cite{midashenglm7b}.
\end{itemize}

This design ensures that audio is no longer treated merely as a text source but as a multi-dimensional channel carrying identity and environmental context, providing high-resolution supervision for cross-modal alignment.

\subsubsection{Video} 
\label{sec:method_dataset_video}
\paragraph{Natural Videos.} 
Existing video parsing systems often exhibit two major blind spots: they underutilize the audio modality (treating it merely as ASR transcripts and ignoring acoustic cues or speaker identity) and they overlook fine-grained camera motion, which is critical for directing attention, shaping narrative rhythm, and conveying emotion. To address these limitations and capture holistic scene dynamics, we curate two natural video datasets: General Video and Camera-aware Video.

\begin{itemize}[topsep=0pt, partopsep=0pt, itemsep=2pt, parsep=0pt, leftmargin=*]
\item \textbf{General Video.} To resolve cross-modal inconsistencies and fully exploit non-linguistic audio cues, we developed a multi-stage curation pipeline to extract fine-grained, time-aligned signals from both modalities, resulting in 511K captioning samples (durations of 5--900s) and 266K parsing samples. This pipeline consists of three steps: \textit{(1) Filtering.} We apply Voice Activity Detection (VAD) to remove silence, and Language Identification (LID) via SenseVoice \cite{an2024funaudiollm} to retain only English and Mandarin content. Clips with severe motion blur or durations shorter than 5 seconds are discarded. \textit{(2) Temporal-Visual Segmentation.} To capture local dynamics, we perform Scene Boundary Detection based on chromatic and luminance variance between adjacent frames. The video is partitioned into semantic segments; clips under 3s or 5 frames are pruned, while those over 60s are sub-segmented. For each segment, we extract visual attributes (atomic actions, object-actor interactions, and basic camera motion) and use Qwen3-VL-235B-A22B to generate detailed event descriptions. \textit{(3) Cross-Modal Synthesis and Structured Representation.} We fuse unimodal streams into a structured hierarchy by calculating the intersection of visual boundaries and audio semantic chunks. At the \textit{segment-level}, the data contains timestamped descriptions paired with visual attributes and audio semantic tags (e.g., ``Instrumental Music'', ``Human Voice''). At the \textit{global-level}, combining Qwen3-VL-235B-A22B with MidasHengLM, we employ dynamic, category-aware schemas to synthesize a holistic video narrative that integrates the full speaker-attributed ASR history. This structured output reduces redundancy and enforces cross-modal consistency, transforming audio from a secondary text source into a primary sensory channel for identity, environment, and narrative context.

\item \textbf{Camera-aware Video.} While the general data captures overarching visual attributes, deep cinematic understanding requires precise spatiotemporal grounding of camera dynamics (e.g., distinguishing fast tracking shots from slow dolly-in shots). To equip MLLMs with this capability, we construct a 191K-sample dataset (4--10\,s duration) sourced from MovieNet \cite{huang2020movienet} and internal datasets. To mitigate the unreliability of automated estimators, we adopt a hybrid pipeline: VGGT \cite{wang2025vggt} provides automated coarse filtering, followed by professional manual annotation of ${\sim}$50K clips. The annotation utilizes a two-level taxonomy adapted from CameraBench \cite{lin2025towards}: L1 categorizes broad motions into \textit{no-motion}, \textit{tracking}, and \textit{non-tracking}; L2 refines \textit{tracking} into 7 subclasses (e.g., pan/tilt/arc tracking) and \textit{non-tracking} into 15 subclasses covering specific translations, rotations, and focal adjustments. Based on these high-precision labels, Qwen3-Omni-30B-A3B synthesizes camera-aware descriptions and temporal parsing boundaries, while Gemini-2.5-Pro generates reasoning-oriented QA pairs. This dedicated corpus significantly enhances the models' temporal precision and capability to integrate complex camera movements into narrative descriptions.
\end{itemize}

\paragraph{Text-Rich Videos.} 
To address the critical need for fine-grained semantic parsing and structured understanding in long-form, text-rich instructional videos, we construct an in-house course video dataset. Specifically, this includes 130K in-depth structured captions, and 79K course video parsing entries. Videos are collected from high-quality educational channels on YouTube, covering university open courses, foundational discipline series, professional skills training, and popular science lectures across multiple domains such as technology, business, humanities, and arts. We use Qwen3-Omni-30B-A3B for data curation, retaining videos with clear objectives, high knowledge density, and rich audio-visual signals.

\begin{itemize}[topsep=0pt, partopsep=0pt, itemsep=2pt, parsep=0pt, leftmargin=*]
\item \textbf{In-Depth Structured Caption.} We generate course reports with a two-step pipeline: (1) Qwen3-Omni-30B-A3B summarizes audio content; (2) Qwen3-VL-235B-A22B fuses the audio summaries with visual content to produce a structured report following a predefined template. The template includes five core dimensions: course title, core abstract, structured outline (via logical sectioning), deep content mining (core concepts, Q\&A, and resources such as code, formulas, and charts), and learning path planning. It significantly enhanced the model's capacity for information distillation and structured summarization in long-form dynamic audio-visual scenarios.

\item \textbf{Parsing.} We build an automated parsing pipeline to convert fragmented data into structured information: (1) keyframe and timestamp extraction using SSIM~ \cite{zhang20252}; (2) fine-grained OCR on keyframes with Qwen3-VL-235B-A22B to accurately extract LaTeX formulas, Markdown tables, format-preserved code blocks, titles, and standard text paragraphs, yielding a series of chronologically ordered frame-level OCR results; (3) semantic-aware temporal aggregation via Qwen3-235B-A22B-Thinking to transform discrete keyframes into continuous semantic segments by dynamically detecting contextual boundaries and intelligently merging evolving content (e.g., stitching scrolling code, updating dynamic tables, and eliminating redundancies); (4) segment-level audio processing using Qwen3-Omni-30B-A3B to extract corresponding audio annotations (including both ASR transcripts and audio captions), integrating these modalities into the localized parsing results; and (5) global audio-visual caption, wherein Qwen3-Omni-30B-A3B first generates a holistic audio description from the full audio track, which is subsequently inputted alongside the video into Qwen3-VL-235B-A22B to yield a comprehensive global description. Ultimately, this pipeline generates a unified JSON dataset formatted as follows: a chronologically ordered series of segments—each comprising a start and end timestamp, consolidated OCR text, and localized audio annotations (available in two modes: ASR or audio caption)—terminating with the overarching global audio-visual description appended at the end. This pipeline effectively transforms long-sequence, complex audio-visual streams into precisely structured, temporally aligned multimodal knowledge.
\end{itemize}

\subsection{Training}
To train Logics-Parsing-Omni as a unified parser that seamlessly integrates localized structural fidelity with high-level semantic interpretation, we initialize our model from Qwen3-Omni-30B-A3B, which has established a robust baseline across a wide range of cross-modal tasks. Instead of training from scratch, we adopt a two-stage progressive training strategy that perfectly mirrors our Omni Parsing framework. Standard autoregressive next-token prediction is used as the supervised fine-tuning (SFT) objective in both stages.

\textbf{Stage 1: Large-scale Multitask Capability \& Knowledge Expansion.}
The core objective of this stage is to construct a ``Panoramic Cognitive Foundation.'' Leveraging a curated large-scale dataset of 16M supervised samples, we perform full-parameter SFT. In this phase, we do not enforce strict category balancing; instead, we prioritize data scale and coverage to build the foundational skills for Holistic Detection (L1) and Fine-grained Recognition (L2):

\begin{itemize}[topsep=0pt, partopsep=0pt, itemsep=2pt, parsep=0pt, leftmargin=*]
\item \textit{Visual Knowledge Injection:} We introduce approximately 12.6 million image-based QA pairs to map visual features into an encyclopedic world knowledge space. This encourages the model to not only ``perceive'' entities but also intuitively associate them with underlying high-level semantic facts, thereby expanding its cognitive breadth.
\item \textit{Atomic Capability Generalization:} We aggregate data from diverse foundational tasks, including document structure parsing, knowledge-aware natural image captioning, image/geometric editing difference captioning, chart/graphics captioning, as well as general audio and audio-visual captioning. This comprehensive exposure ensures the model masters pure structural parsing and basic semantic descriptions before being overwhelmed by complex reasoning tasks.
\end{itemize}

Following Stage~1, the model accumulates extensive knowledge reserves and preliminary cross-task generalization capabilities. However, due to the sheer volume of image QA data, the model exhibits certain task distribution biases and has yet to fully master the highly structured logical reasoning required by our paradigm. To unlock its precision potential, we proceed to the second stage.

\textbf{Stage 2: Unified Parsing Alignment \& Balanced Instruction Tuning.} 
To rectify the task distribution bias inherited from Stage~1 and achieve deep integration of perception and cognition (i.e., L3 Semantic Interpretation), we shift our training paradigm from scale-driven knowledge accumulation to precise, balanced instruction tuning. By constructing a high-quality dataset comprising 5M samples, we explicitly prioritize instruction density, data quality, and structural diversity:

\begin{itemize}[topsep=0pt, partopsep=0pt, itemsep=2pt, parsep=0pt, leftmargin=*]
\item \textit{Multi-domain Task Reinforcement:} The cornerstone of this stage is the high-quality data synthesized by our Omni Parsing Framework (as illustrated in Figure~\ref{fig:data_synthesis_pipeline}). This activates the full L1-to-L3 progressive parsing pipeline across all domains detailed in our corpus. Specifically, it includes: document parsing (covering pure structural and joint structure-semantic extraction); image parsing (encompassing natural images, knowledge-aware images, and graphics, with tasks spanning single-image and multi-image difference parsing); general audio parsing; and comprehensive video parsing (spanning natural, camera-aware, and text-rich videos). To preserve the model's general conversational capabilities, we also integrate a minor, meticulously balanced proportion of multi-modal captioning and QA data.

\item \textit{Omni-modal Schema Alignment:} Through fine-tuning on this strictly balanced distribution, we achieve a dual alignment of structured extraction and free-form semantic interaction. For omni-parsing tasks, we explicitly enforce a unified output schema, training the model to map heterogeneous inputs---whether documents, static images, or dynamic audio-visual streams---into standardized JSON formats that seamlessly bridge spatial-temporal grounding with semantic reasoning. Concurrently, it maintains fluent natural language generation for general QA and captioning. Consequently, the model learns to process omni-modal signals under a rigorous progressive parsing paradigm without sacrificing its conversational versatility.
\end{itemize}

Ultimately, this two-stage methodology ensures that Logics-Parsing-Omni not only inherits vast encyclopedic knowledge but also demonstrates highly consistent instruction-following capabilities. It successfully realizes the efficient synergy between fine-grained pixel/signal-level perception and high-level semantic cognition, achieving the optimal state designed by our framework.

\textbf{Experimental settings.} We implemented our framework using Megatron-SWIFT \cite{zhao2024swiftascalablelightweightinfrastructure} across both training stages. For model architecture, we unfroze all components of Qwen3-Omni-30B-A3B except the talker, including all parameters of the LLM, vision encoder, audio encoder, and aligners. For multimodal data processing, we established precise input specifications for distinct modalities: images retain their original aspect ratio and resolution; for videos, we applied a uniform sampling rate of 2.0 FPS, set the maximum number of extracted frames to 768, and utilized the original resolution for individual frames. To further enhance computational efficiency and ensure load balancing during multimodal long-sequence training, we introduced a sequence packing strategy, maintaining a constant global batch size of 32 and extending the maximum sequence length to 56k to fully accommodate the contextual requirements of long documents and audio-visual inputs. The optimization process employed a base learning rate of $1\text{e-}5$, coupled with a warmup ratio of 0.05 and a cosine decay strategy, to achieve smooth convergence during model training.

\section{Evaluation} 
\subsection{OmniParsingBench}
\subsubsection{Overall} 
To rigorously evaluate the unified parsing capabilities of our model across diverse modalities, we construct OmniParsingBench, a comprehensive, large-scale, and high-quality evaluation corpus. Unlike traditional single-task benchmarks, OmniParsingBench is designed to assess the full spectrum of parsing performance—from fundamental signal detection to complex semantic reasoning—across six primary domains: Document, Natural Image, Graphics, Audio, Natural Video and Text-Rich Video.

Our evaluation framework strictly aligns with the proposed three-stage architecture (L1–L3), systematically assessing performance from L1-Holistic Detection (spatio-temporal grounding and classification), L2-Fine-grained Recognition (symbol extraction, attribute identification, and structural recovery), to L3-Multi-level Interpreting (semantic consistency and hallucination resistance). To provide a concise view of model capabilities, we aggregate these fine-grained metrics into two core scores: Perception, which evaluates signal precision and structural fidelity (dominating L1 and L2), and Cognition, which evaluates logical reasoning and semantic understanding (dominating L3).

As detailed in Table~\ref{tab:summary_results}, Logics-Parsing-Omni demonstrates highly competitive or state-of-the-art capabilities across all six diverse modalities when evaluated on Overall (Ovr.), Perception (Perc.), and Cognition (Cog.) metrics. Notably, our model consistently surpasses all evaluated baselines—including the leading proprietary Gemini-3-Pro—in the Overall and Cognition metrics of the Graphics, Audio, and Text-Rich Video domains.  This superiority is particularly pronounced in the Cognition metric, where Logics-Parsing-Omni exhibits exceptional logical reasoning and semantic understanding, achieving top-tier scores such as 92.12 in Graphics and 80.85 in Text-Rich Video. While Gemini-3-Pro maintains an advantage in the fundamental perception of Natural Images, Graphics, Audio, and Documents, as well as a marginal lead in Natural Video, our model significantly outperforms other open-weight counterparts (e.g., the Qwen series) in nearly all metrics. These quantitative results validate the efficacy of our L1–L3 architecture, demonstrating that Logics-Parsing-Omni successfully bridges fundamental signal detection with complex multi-modal interpreting.

\begin{table*}[htbp]
\centering
\caption{Performance comparison of various models on OmniParsingBench.}
\label{tab:summary_results}
\renewcommand{\arraystretch}{1.3} 
\resizebox{\linewidth}{!}{%
\begin{threeparttable}
    \begin{tabular}{l *{16}{c}} 
        \toprule
        
        \multirow{2}{*}{\textbf{Model}} & 
        \multicolumn{3}{c}{\textbf{Natural Image}} & 
        \multicolumn{3}{c}{\textbf{Graphics}} & 
        \multicolumn{1}{c}{\textbf{Document}} & 
        \multicolumn{3}{c}{\textbf{Audio}} & 
        \multicolumn{3}{c}{\textbf{Natural Video}} & 
        \multicolumn{3}{c}{\textbf{Text-Rich Video}} \\
        
        \cmidrule(lr){2-4} \cmidrule(lr){5-7} \cmidrule(lr){8-8} \cmidrule(lr){9-11} \cmidrule(lr){12-14} \cmidrule(lr){15-17}
        
        & \textbf{Ovr.} & \textbf{Perc.} & \textbf{Cog.} 
        & \textbf{Ovr.} & \textbf{Perc.} & \textbf{Cog.} 
        & \textbf{Perc.} 
        & \textbf{Ovr.} & \textbf{Perc.} & \textbf{Cog.} 
        & \textbf{Ovr.} & \textbf{Perc.} & \textbf{Cog.} 
        & \textbf{Ovr.} & \textbf{Perc.} & \textbf{Cog.} \\
        
        \midrule
        
        Gemini-3-Pro 
        & \textbf{61.20} & 55.96 & \textbf{66.44} 
        & \underline{87.03} & \textbf{84.21} & 87.43 
        & \textbf{87.01} 
        & \underline{79.40} & \textbf{72.90} & 85.89
        & \textbf{63.40} & \textbf{57.87} & \textbf{68.92} 
        & \underline{64.37} & \textbf{58.54} & \underline{70.20} \\
        
        GPT-5.2 
        & 39.94 & 37.77 & 42.12 
        & 82.71 & 69.86 & \underline{91.48} 
        & 77.43 
        & -- & -- & -- 
        & -- & -- & -- 
        & -- & -- & -- \\
        
        Qwen3.5-397B-A17B 
        & 57.40 & \textbf{56.95} & 57.85 
        & 82.81 & 73.77 & 83.13
        & 81.09 
        & -- & -- & -- 
        & -- & -- & -- 
        & -- & -- & -- \\
        
        Qwen3-VL-235B-A22B 
        & 58.61 & \underline{56.23} & 60.99 
        & 79.49 & 71.51 & 83.46
        & 84.47 
        & -- & -- & -- 
        & -- & -- & -- 
        & -- & -- & -- \\
        
        Qwen3-VL-30B-A3B 
        & 50.92 & 48.91 & 52.94 
        & 73.25 & 65.71 & 79.36 
        & 78.94 
        & -- & -- & -- 
        & -- & -- & -- 
        & -- & -- & -- \\
        
        Qwen3-Omni-30B-A3B 
        & 47.36 & 46.85 & 47.88 
        & 77.46 & 70.75 & 78.25 
        & 73.50 
        & 75.17 & 62.13 & \underline{88.22} 
        & 45.23 & 34.15 & 56.32
        & 26.86 & 10.22 & 43.50 \\
        
        Logics-Parsing-Omni(Ours) 
        & \underline{59.07} & 53.77 & \underline{64.37} 
        & \textbf{88.66} & \underline{82.01} & \textbf{92.12} 
        & \underline{84.90} 
        & \textbf{79.63} & \underline{69.27} & \textbf{89.99} 
        & \underline{61.12} & \underline{56.09} & \underline{66.15} 
        & \textbf{69.12} & \underline{57.39} & \textbf{80.85} \\
        
        \bottomrule
    \end{tabular}%
\begin{tablenotes}
    \fontsize{14}{14}\selectfont 
    \item[] \textit{Note:} Bold text indicates the best result, and underlined text indicates the second-best result.
\end{tablenotes}
\end{threeparttable}
}
\end{table*}

\subsubsection{Image} 
\paragraph{Natural Images.} 

To rigorously assess the perceptual and cognitive capabilities of models on natural images, we utilize the Natural Image module of OmniParsingBench. This subset comprises 1,000 diverse natural images alongside 1,000 images annotated with authoritative domain knowledge (covering domains that require knowledge-aware parsing, such as landmark buildings, brands, and biological species). To ensure structural compatibility across heterogeneous model architectures, all models must output their parsing results in a predefined structured format. Comprehensive details regarding dataset curation, evaluation protocols, metric formulations, and extended results are provided in Appendix~\ref{subsec:appendix_logics_image_parsing}.

The evaluation metrics are systematically structured along two core dimensions:
\begin{itemize}[topsep=0pt, partopsep=0pt, itemsep=2pt, parsep=0pt, leftmargin=*]
\item \textbf{Perception:} This aspect assesses the model's fundamental signal detection and symbol extraction capabilities across two sequential stages:
(1) \textit{Location}: Quantifies spatial grounding ability by calculating the mean localization recall (IoU \textgreater 0.5) for general entities, knowledge-aware entities, and text blocks.
(2) \textit{Content}: Applies rigorous exact-match and semantic verification to successfully localized instances. This includes evaluating the semantic fidelity of general entities, ensuring exact OCR and language-type matching for text, and verifying that knowledge-aware entities accurately reference their authoritative identifiers.
\item \textbf{Cognition:} This aspect evaluates the logical coherence, depth, and hallucination resistance of the global captions derived from the parsed structures. It is subdivided into:
(1) \textit{General Caption QA}: Following the CaptionQA protocol, an LLM evaluates the global description from image parsing across nine multidimensional semantic aspects (e.g., theme, spatial relationships, textual content, visual style) using grounded question-answer pairs.
(2) \textit{Knowledge Reference}: For images containing knowledge-aware entities, this metric computes the proportion of captions that explicitly and accurately integrate all verified knowledge identifiers into the global narrative, thereby ensuring factual correctness and structural-semantic alignment.
\end{itemize}

\begin{table}[htbp]
\centering
\caption{Evaluation results on the Natural Image module of OmniParsingBench.}
\label{tab:image_parsing_eval_result}
\renewcommand{\arraystretch}{1.2} 
\resizebox{\textwidth}{!}{
\begin{tabular}{l c ccccc ccc}
\toprule
\multirow{3}{*}{\textbf{Model}} 
& \multirow{3}{*}{\textbf{Overall}$^{\P}$}
& \multicolumn{5}{c}{\textbf{Perception}}
& \multicolumn{3}{c}{\textbf{Cognition}} \\
\cmidrule(lr){3-7} \cmidrule(lr){8-10}
& & \multirow{2}{*}{Location}
& \multicolumn{3}{c}{Content$^{*}$} 
& \multirow{2}{*}{\textbf{Avg.}}
& \multirow{2}{*}{General} 
& \multirow{2}{*}{Knowledge} 
& \multirow{2}{*}{\textbf{Avg.}} \\
\cmidrule(lr){4-6}
& & & Entity & Knowledge & Text & & & & \\
\midrule
Gemini-3-Pro\footnotemark[13] & \textbf{61.20} & 74.33 & 97.35 & \textbf{59.44} & \textbf{70.51} & 55.96 & \underline{74.14} & \textbf{58.74} & \textbf{66.44} \\
GPT-5.2$^{\dag}$\footnotemark[9] & 39.94 & 52.91 & 93.08 & -- & 44.78 & 37.77 & 66.46 & 17.77 & 42.12 \\
Qwen3.5-397B-A17B\footnotemark[16] & 57.40 & \textbf{81.98} & \underline{97.60} & 51.80 & \underline{63.56} & \textbf{56.95} & 66.41 & 49.28 & 57.85 \\
Qwen3-VL-235B-A22B~\cite{Qwen3-VL} & 58.61 & 77.99 & 97.41 & \underline{58.68} & 57.89 & \underline{56.23} & 68.26 & \underline{53.71} & 60.99 \\
Qwen3-VL-30B-A3B~\cite{Qwen3-VL} & 50.92 & 78.09 & \textbf{97.65} & 44.77 & 54.40 & 48.91 & 63.75 & 42.12 & 52.94 \\
Qwen3-Omni-30B-A3B~\cite{Qwen3-Omni} & 47.36 & 75.78 & 97.15 & 33.02 & 60.40 & 46.85 & 61.65 & 34.10 & 47.88 \\
\midrule
Logics-Parsing-Omni (Ours) & \underline{59.07} & \underline{79.21} & 97.55 & 48.99 & 62.14 & 53.77 & \textbf{82.60} & 46.13 & \underline{64.37} \\
\bottomrule
\multicolumn{10}{l}{$^*$ Content metrics are semantic accuracy computed conditionally on successfully localized instances.} \\
\multicolumn{10}{l}{$^\P$ Overall Score = mean(Perception Avg, Cognition Avg.).} \\
\multicolumn{10}{l}{$^\dag$ GPT-5.2 exhibits policy-induced constraints in knowledge identification; Knowledge Score is excluded in Avg. Score.} \\
\end{tabular}}
\end{table}

\noindent \textit{Result Analysis.}
As shown in Table~\ref{tab:image_parsing_eval_result}, \textbf{Logics-Parsing-Omni} achieves an Overall score of 59.07\%, outperforming all evaluated open-weight models (e.g., Qwen3.5-397B-A17B at 57.40\%) and demonstrating a highly competitive balance between structural fidelity and semantic understanding.

For the Perception (Avg. 53.77\%), our model demonstrates strong signal detection capabilities, achieving a highly competitive Location score of 79.21\%, while maintaining robust general entity and text recognition.

For the Cognition (Avg. 64.37\%), Logics-Parsing-Omni exhibits exceptional semantic depth. Notably, it achieves a state-of-the-art General Caption score of 82.60\%, surpassing even the proprietary Gemini-3-Pro (74.14\%). This confirms our framework produces logically coherent and factual narratives (Knowledge Ref. 46.13\%) driven by explicit structured parsing rather than hallucination.

\paragraph{Graphics.} 
To rigorously evaluate the perceptual and cognitive capabilities of models on information-dense visual elements, we conduct our evaluation on the Graphics module of OmniParsingBench. This subset encompasses two highly structured domains requiring complex spatial and logical deduction: Charts and Geometric Figures. Consistent with our overall evaluation framework, we specifically instantiate the Perception and Cognition dimensions for the Graphics module as follows:

\begin{itemize}[topsep=0pt, partopsep=0pt, itemsep=2pt, parsep=0pt, leftmargin=*]
\item \textbf{Perception:} This dimension evaluates fundamental visual decomposition and exact structural recovery. It is subdivided into: 
    (1) \textit{Location}: Assesses precise geometric spatial positioning (e.g., exact coordinates and radii). 
    (2) \textit{Content - Element}: Evaluates the extraction of fundamental geometric primitives (points, lines, circles).
    (3) \textit{Content - Relation}: Measures the extraction of explicit topological and quantitative relationships in geometry.
    (4) \textit{Content - Chart}: Evaluates structural reverse-rendering exactness, measuring data extraction in statistical charts (via F1-score) and topological correctness in structural diagrams (via Mermaid similarity).
\item \textbf{Cognition:} This dimension assesses the model's ability to bind visual structures with mathematical or factual logic. Following the CaptionQA protocol used in Natural Images, an LLM evaluates semantic coverage and reasoning capabilities using grounded question-answer pairs. It is subdivided into: 
    (1) \textit{Element}: Evaluates fundamental information extraction across both domains (e.g., basic geometric elements and chart OCR data). 
    (2) \textit{Relation}: Captures the ability to comprehend and bind complex topological and visual structures. 
    (3) \textit{Reasoning}: Represents complex logical deduction capability, specifically the ability to analyze implicit data trends, fluctuations, and perform numerical reasoning in charts.
\end{itemize}

\begin{table*}[htbp]
    \centering
    \caption{Evaluation results on the Graphics Module of OmniParsingBench.}
    \label{tab:Graphics_exp}
    \renewcommand{\arraystretch}{1.2}
    \setlength{\tabcolsep}{5pt}
    \resizebox{\textwidth}{!}{%
    \begin{tabular}{l c ccccc cccc}
        \toprule
        \multirow{3}{*}{\textbf{Model}} & \multirow{3}{*}{\textbf{Overall}} & \multicolumn{5}{c}{\textbf{Perception}} & \multicolumn{4}{c}{\textbf{Cognition}} \\
        \cmidrule(lr){3-7} \cmidrule(lr){8-11}
        & & \multirow{2}{*}{Location} & \multicolumn{3}{c}{Content} & \multirow{2}{*}{\textbf{Avg.}} & \multirow{2}{*}{Element} & \multirow{2}{*}{Relation} & \multirow{2}{*}{Reasoning} & \multirow{2}{*}{\textbf{Avg.}} \\
        \cmidrule(lr){4-6}
        & & & Element & Relation & Chart & & & & & \\
        \midrule
        Gemini-3-Pro       & \underline{87.03} & \underline{78.87} & \textbf{86.21} & \textbf{83.95} & 87.80          & \textbf{84.21} & 89.37          & \underline{89.83} & 83.11          & 87.43          \\
        GPT-5.2            & 82.71             & 45.62             & \underline{83.30} & 66.97       & 83.56          & 69.86          & \underline{93.18} & 87.30          & \textbf{93.95} & \underline{91.48} \\
        Qwen3.5-397B-A17B  & 82.81             & 46.48             & 82.15          & \underline{74.55} & \textbf{91.91} & 73.77       & 86.21          & 89.22          & 73.97          & 83.13          \\
        Qwen3-VL-235B-A22B & 79.49             & 52.56             & 83.20          & 66.28          & 83.99          & 71.51          & 85.68          & 82.97          & 81.74          & 83.46          \\
        Qwen3-VL-30B-A3B   & 73.25             & 62.03             & 80.89          & 52.54          & 67.40          & 65.71          & 83.95          & 79.71          & 74.43          & 79.36          \\
        Qwen3-Omni-30B-A3B & 77.46             & 59.94             & 81.79          & 55.06          & 86.19          & 70.75          & 81.25          & 79.53          & 73.97          & 78.25          \\
        \midrule
        \textbf{Logics-Parsing-Omni (Ours)} & \textbf{88.66} & \textbf{85.46} & 82.32 & 72.15 & \underline{88.12} & \underline{82.01} & \textbf{96.57} & \textbf{91.65} & \underline{88.13} & \textbf{92.12} \\
        \bottomrule
    \end{tabular}
    }
\end{table*}

\textit{Result Analysis.}
As shown in Table~\ref{tab:Graphics_exp}, Logics-Parsing-Omni achieves an overall accuracy of 87.43, showing a significant improvement over the baseline Qwen3-Omni-30B-A3B (74.93) and demonstrating strong competitiveness against state-of-the-art models. For the Perception, our model attains an exceptional score of 86.30 in fine-grained Location, significantly outperforming Gemini-3-Pro and GPT-5.2.

Anchored by this precise structural perception, Logics-Parsing-Omni dominates the Cognition, achieving the highest average accuracy of 92.19. Notably, our model excels in fundamental extraction and structural binding, achieving state-of-the-art scores in both the Element (96.16) and Relation (92.27) metrics. Furthermore, in the complex Reasoning metric—which evaluates logical deduction capabilities—our model reaches a highly competitive 88.13, surpassing Gemini-3-Pro (83.11) and closely following GPT-5.2 (93.95).

These results validate our proposed progressive parsing paradigm: the fine-grained structural fidelity helps mitigate the limitations of superficial image captioning, laying a robust foundation for rigorous spatial deduction and complex logical reasoning. Detailed evaluation dimensions and results are provided in Appendix~\ref{sec:appendix_graphics}.

\subsubsection{Document} 
In this subsection, we report the experimental results of our Logics-Parsing-Omni model on the proposed LogicsDocBench~\footnotemark[14], which effectively validates the effectiveness of the proposed omni model on the document parsing task. To be specific, LogicsDocBench contains 900 carefully curated page-level PDF images covering both traditional Parsing-1.0 tasks and newly introduced Parsing-2.0 scenarios, including flowcharts, music sheets and pseudo-code blocks. Since our omni model does not include Parsing-2.0 scenarios training, we only report the evaluation results on Parsing-1.0 tasks. Quantitative comparison results are summarized in Table ~\ref{tab:logicsdocbench_results}.

\begin{table}[h]
\footnotesize
\caption{Comprehensive evaluation of document reading on LogicsDocBench.}
\label{tab:logicsdocbench_results}
\centering
\setlength{\tabcolsep}{1.6pt}
\resizebox{\linewidth}{!}{%
\begin{tabular}{lcccccc}
\toprule[.9pt]
\multirow{2}{*}{\textbf{Model}} 
  & \multirow{3}{*}{\makecell{\textbf{Overall}}}
  & \multicolumn{5}{c}{\textbf{Perception}} \\
\cmidrule(lr){3-7}
  & 
  & Text$^{Edit}$ $\downarrow$  
  & Formula$^{CDM}$  
  & Table$^{TEDs}$  
  & Table$^{TEDS_s}$  
  & Reading Order$^{Edit}$ $\downarrow$ \\  
\midrule
\multicolumn{7}{c}{\textbf{Pipeline}} \\
\midrule
  MinerU2-pipeline~\cite{wang2024mineru} & 77.93 & 0.1013 & 75.89 & 68.03 & 76.14 & 0.2157  \\ 
  MinerU2.5~\cite{niu2025mineru2} & \underline{86.30} & 0.0754 & \textbf{85.07} & 81.36 & 84.48 & 0.1334 \\ 
  PP-StructureV3~\cite{cui2025paddleocr30technicalreport} & 70.26 & 0.1155 & 78.20 & 44.12 & 74.65 & 0.1891 \\
  PaddleOCR-VL~\cite{cui2025paddleocrvlboostingmultilingualdocument} & 85.09 & \textbf{0.0581} & 83.80 & 77.27 & 81.60 & 0.1404 \\
  PaddleOCR-VL-1.5~\cite{cui2026paddleocr} & 85.43 & \underline{0.0587} & 82.78 & 79.38 & 84.30 & \underline{0.1212}  \\
  MonkeyOCR-pro-1.2B~\cite{li2025monkeyocrdocumentparsingstructurerecognitionrelation} & 77.57 & 0.1653 & 79.85 & 69.39 & 74.77 & 0.2788 \\ 
  GLM-OCR~\footnotemark[15] & 85.59 & 0.0689 & 84.10 & 79.57 & 83.90 & 0.1490  \\
\midrule
\multicolumn{7}{c}{\textbf{End-to-end Model}} \\
\midrule
  Qwen3-Omni-30B-A3B~\cite{Qwen3-Omni} & 73.50 & 0.2093 & 72.07 & 69.37 & 73.49 & 0.2317 \\
  HunyuanOCR~\cite{team2025hunyuanocr} & 76.11 & 0.1504 & 71.62 & 71.75 & 76.47 & 0.2036  \\
  DeepSeek-OCR~\cite{wei2025deepseek} & 74.45 & 0.1650 & 76.90 & 62.94 & 66.76 & 0.2189  \\
  DeepSeek-OCR 2~\cite{wei2026deepseek} & 77.04 & 0.1111 & 76.92 & 65.32 & 70.48 & 0.1554  \\   
  InternVL35-241B~\cite{wang2025internvl3} & 76.03 & 0.2035 & 83.23 & 65.21 & 72.00 & 0.2138 \\ 
  GPT-5.2~\footnotemark[9] & 77.43 & 0.2072 & 83.41 & 69.61 & 74.60 & 0.2148  \\ 
  Qwen3-VL-30B-A3B~\cite{Qwen3-VL} & 78.94 & 0.1340 & 81.52 & 68.69 & 75.33 & 0.1718  \\
  Qwen3.5-397B-A17B~\footnotemark[16] & 81.09 & 0.0868 & 80.37 & 71.57 & 78.48 & 0.1450 \\
  Gemini-2.5-Pro~\cite{comanici2025gemini} & 84.29 & 0.0985 & 81.07 & \underline{81.64} & \underline{87.67} & 0.1689  \\
  dots.ocr~\footnotemark[6] & 84.51 & 0.0629 & 81.60 & 78.21 & 82.49 & \textbf{0.1153}  \\
  Qwen3-VL-235B-A22B~\cite{Qwen3-VL} & 84.47 & 0.0817 & 84.00 & 77.58 & 81.92 & 0.1405 \\
  Gemini-3-Pro~\footnotemark[13] & \textbf{87.01} & 0.0815 & \underline{84.86} & \textbf{84.33} & \textbf{90.66} & 0.1468 \\
\midrule 
  Logics-Parsing-Omni (Ours) & 84.9 & 0.0666 & 80.41 & 80.93 & 84.25 & 0.1223 \\ 
\bottomrule[.9pt]
\end{tabular}
}%
\end{table}

\textit{Result Analysis.}
From the experimental results presented in Table ~\ref{tab:logicsdocbench_results}, we observe that our proposed Logics-Parsing-Omni model outperforms all Qwen-series models by a large margin, including Qwen3-Omni-30B-A3B, Qwen3-VL-30B-A3B and the newly released Qwen3.5-397B-A17B, which highlights its superior capability in complex document parsing. Although slightly inferior to Gemini-3-Pro, Logics-Parsing-Omni surpasses all other general-purpose vision-language large models (VLMs), as well as some of the specialized OCR-focused models, such as dots.ocr, DeepSeek-OCR series and so on. These experimental results provide strong evidence of the effectiveness and robustness of Logics-Parsing-Omni as a generalized VLM in handling challenging, real-world document parsing tasks. Furthermore, specific visual examples of the evaluated cases are provided in the Appendix (see Fig.~\ref{fig:img_doc_geo_case} and Fig.~\ref{fig:img_doc_chart_case}).

~\footnotetext[1]{\url{https://doc2x.noedgeai.com/}}
~\footnotetext[2]{\url{https://www.textin.com}}
~\footnotetext[3]{\url{https://mathpix.com/}}
~\footnotetext[4]{\url{https://github.com/datalab-to/marker}}
~\footnotetext[5]{\url{https://github.com/breezedeus/Pix2Text}}
~\footnotetext[6]{\url{https://github.com/rednote-hilab/dots.ocr}}
~\footnotetext[7]{\url{https://ocrflux.pdfparser.io/}}
~\footnotetext[8]{\url{https://seed.bytedance.com/zh/seed1_6}}
~\footnotetext[9]{\url{https://openai.com/zh-Hans-CN/index/introducing-gpt-5/}}
~\footnotetext[10]{\url{https://mistral.ai/news/mistral-ocr}}
~\footnotetext[11]{\url{https://github.com/lavvsharma/ocr_verse}}
~\footnotetext[12]{\url{https://openai.com/zh-Hans-CN/index/hello-gpt-4o/}}
~\footnotetext[13]{\url{https://aistudio.google.com/models/gemini-3}}
~\footnotetext[14]{\url{https://github.com/alibaba/Logics-Parsing}}
~\footnotetext[15]{\url{https://github.com/zai-org/GLM-OCR}}
~\footnotetext[16]{\url{https://github.com/QwenLM/Qwen3.5}}

\subsubsection{Audio} 
To rigorously evaluate the capabilities of models in fine-grained parsing and comprehensive understanding within audio-only scenarios, we introduce the Audio Module of OmniParsingBench. The benchmark comprises 2 main tasks (perception and cognition) and 5 detailed task types, encompassing a total of 1014 cases.

\begin{itemize}[topsep=0pt, partopsep=0pt, itemsep=2pt, parsep=0pt, leftmargin=*]
\item \textbf{Perception.} To evaluate the model's ability to identify audio categories and their precise temporal boundaries.
\begin{itemize}[topsep=0pt, partopsep=0pt, itemsep=2pt, parsep=0pt, leftmargin=*]
\item \textbf{Data Construction.} 
(1) \textit{Audio Event Detection} cases were stochastically sampled from the AudioSet-strong \cite{hershey2021benefittemporallystronglabelsaudio} dataset. For evaluation clarity, we re-categorized the labels into six primary classes: Human Voice, Instrumental Music, Environmental Noise, Traffic, Animals, and Daily Life Sounds. This task is further bifurcated into \textit{Identity Recognition} and \textit{Temporal Localization}.
(2) \textit{ASR} samples were randomly selected from the Librispeech \cite{panayotov2015librispeech} corpus to serve as the ground truth for speech transcription.
\item \textbf{Evaluation Metrics.} The perception task utilizes Key F1 and Time F1 metrics, calculated using a Macro Average across two dimensions. 
(1) \textit{Key-level Metrics}: To assess category-level performance, we calculate Key F1 by balancing Key Recall (the fraction of ground truth keys detected) and Key Precision (the fraction of predicted keys that are correct).
(2) \textit{Time-level Metrics}: Time-based F1 score is derived from Time Recall and Time Precision, which measure the temporal overlap between predicted segments and ground truth intervals relative to their total durations. 
(3) \textit{ASR WER}: It is employed to measure the linguistic accuracy of the model's transcriptions against the ground truth.
\end{itemize}
\item \textbf{Cognition.} To assess the model's high-level semantic understanding and reasoning capabilities regarding audio content.
\begin{itemize}[topsep=0pt, partopsep=0pt, itemsep=2pt, parsep=0pt, leftmargin=*]
\item \textbf{Data Construction.} The cognition set is a hybrid corpus consisting of the Worldsense \cite{hong2025worldsense} dataset and a self-constructed dataset. The latter was generated following the pipeline described in~\ref{sec:method_dataset_audio}, where Gemini-3-flash-preview processed segmented audio to generate complex Question-Answering (QA) pairs.
\item \textbf{Subtasks.} The cognition task focuses on two core dimensions. 
(1) \textit{Audio Information Extraction}: The ability to distill specific information and key details from complex audio environments.
(2) \textit{Audio Recognition}: Evaluating the model’s deeper comprehension and logical interpretation of audio events.
\end{itemize}
\end{itemize}

\begin{table*}[htbp]
\centering
\begin{threeparttable} 
\caption{Evaluation results on the Audio Module of OmniParsingBench}
\label{tab:logicsaudiobench}
\renewcommand{\arraystretch}{1.3}
\fontsize{7.5}{7.5}\selectfont 
\begin{tabular}{l c cccc ccc}
\toprule
\multirow{2}{*}{\textbf{Model}} & \multirow{2}{*}{\textbf{Overall}} & \multicolumn{4}{c}{\textbf{Perception}} & \multicolumn{3}{c}{\textbf{Cognition}} \\
\cmidrule(lr){3-6} \cmidrule(lr){7-9}
& & Time & Key & ASR$\downarrow$ & \textbf{Avg.} & Info. Ext. & Rec. & \textbf{Avg.}   \\
\midrule
Gemini-3-Pro & \underline{79.4} & \textbf{64.89} & \textbf{76.62} & \underline{0.2280} & \textbf{72.90} & 94.29 & 77.49 & 85.89 \\
Qwen3-Omni-30B-A3B & 75.17 & 59.12 & 65.39 & 0.3813 & 62.13 & \textbf{95.92} & \underline{80.52} & \underline{88.22} \\
Logics-Parsing-Omni(Ours) & \textbf{79.63} & \underline{59.81} & \underline{65.48} & \textbf{0.1747} & \underline{69.27} & \underline{94.69} & \textbf{85.28} & \textbf{89.99} \\
\bottomrule
\end{tabular}
\begin{tablenotes}
\fontsize{8}{8}\selectfont 
\item[] \textit{Note:} Bold text indicates the best result, and underlined text indicates the second-best result.
\end{tablenotes}
\end{threeparttable}
\end{table*}

\begin{figure}[htbp]
    \centering
    \includegraphics[width=1.0\textwidth]{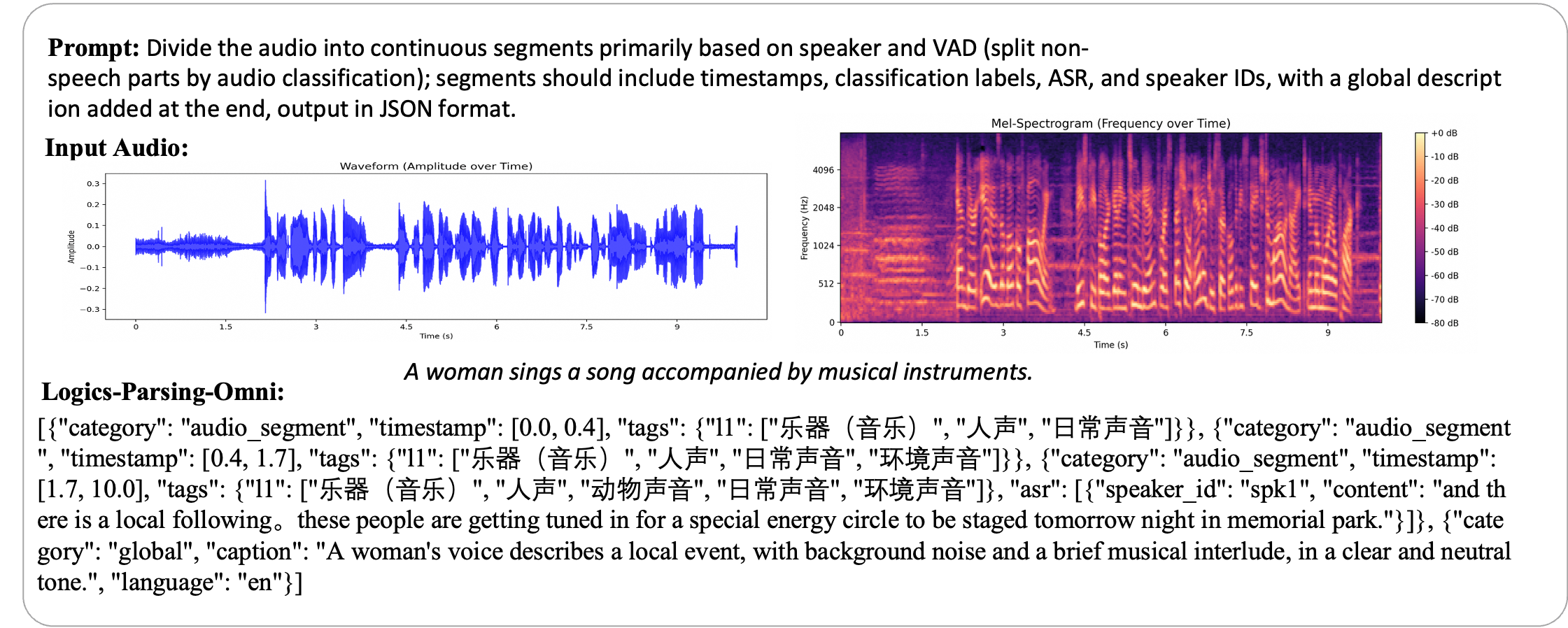}
    \caption{Qualitative examples illustrating the comprehensive audio parsing capability of Logics-Parsing-Omni.}
    \label{fig:audio_show_case}
\end{figure}

\textit{Result Analysis.}
As summarized in Table~\ref{tab:logicsaudiobench}, the overall score is aggregated from Perception and Cognition dimensions. The exact calculation rules are detailed in Appendix~\ref{sec:appendix_audios}. Logics-Parsing-Omni establishes a new state-of-the-art (SOTA) on the Audio Module of OmniParsingBench with an overall score of 79.63. In the cognition tasks, it achieves 94.69\% accuracy in audio information extraction and 85.28\% in audio recognition—surpassing the strongest baseline by a significant 4.76\% margin—which underscores its exceptional semantic comprehension. Qualitatively (Figure~\ref{fig:audio_show_case}), the model demonstrates robust parsing capabilities by generating structured JSON outputs with high-precision timestamps and tags, complemented by descriptive captions that effectively capture the dynamic progression of the auditory scene.

\subsubsection{Video} 
\paragraph{Natural Videos.}
We construct the Natural Video module of OmniParsingBench, which includes 601 samples covering perception and cognition (sourced from the Worldsense dataset and newly curated YouTube multi-turn QA) as well as 520 strictly controlled single-motion scenes, all of which have been meticulously annotated by humans. The comprehensive evaluation results are summarized in Table~\ref{tab:logics_natural_videos_combined}. The exact calculation rules are detailed in Appendix~\ref{sec:appendix_videos}. We evaluate these scenarios across two core dimensions:

\begin{itemize}[topsep=4pt, partopsep=0pt, itemsep=4pt, parsep=0pt, leftmargin=*]
\item \textbf{Perception:} 
This dimension evaluates the model's accuracy in capturing low-level audio-visual details and recognizing technical terminology. The evaluation covers three core aspects:
\begin{itemize}[topsep=2pt, partopsep=0pt, itemsep=2pt, parsep=0pt, leftmargin=*]
    \item \textit{Audio-centric parsing:} Assesses identity recognition and speech-to-text accuracy (ASR WER).
    \item \textit{Temporal Grounding:} Evaluates the precision of audio-visual temporal localization and ordered video event detection, which requires the model to accurately identify events while maintaining their correct chronological sequence.
    \item \textit{Camera Motion Analysis:} Measures the classification accuracy of professional camera terminology (L1/L2 motions) and their joint prediction performance.
\end{itemize}

\item \textbf{Cognition:} 
This dimension moves beyond basic parsing to assess complex reasoning capabilities and the natural integration of multi-modal concepts into language. The evaluation is categorized into three cross-modal reasoning aspects:
\begin{itemize}[topsep=2pt, partopsep=0pt, itemsep=2pt, parsep=0pt, leftmargin=*]
    \item \textit{Audio Reasoning:} Evaluates auditory semantic understanding, including acoustic event counting, semantic sound recognition, and complex audio information extraction.
    \item \textit{Visual Semantics \& Dynamics:} Assesses visual-centric comprehension, such as attribute recognition and event description. Crucially, it also evaluates the natural integration of spatial-dynamic concepts (i.e., camera motion) into global video captions. To ensure robust evaluation of these dynamic descriptions, we adopt an LLM-as-a-Judge paradigm (Qwen3-235B-A3B) to validate semantic consistency by mapping descriptions into eight macro motion concepts (e.g., zoom in/out, pan, track), rather than relying on strict lexical matching.
    \item \textit{Audio-Visual Reasoning:} Tests the model's ability to synthesize information across modalities, focusing on complex tasks like audio-visual emotion change analysis and causal reasoning (i.e., understanding the underlying physical relationship between specific sounds and visual actions).
\end{itemize}
\end{itemize}


\begin{table*}[htbp]
    \centering
    \caption{Evaluation results on the Natural Video Module of OmniParsingBench, integrating both General Videos and Camera-aware Videos metrics.}
    \label{tab:logics_natural_videos_combined}
    \renewcommand{\arraystretch}{1.3} 
    \resizebox{\textwidth}{!}{
        \begin{threeparttable}
            \begin{tabular}{l c cccccc ccccccccc}
                \toprule
                \multirow{3}{*}{\textbf{Model}} & 
                \multirow{3}{*}{\textbf{Overall}} & 
                \multicolumn{6}{c}{\textbf{Perception}} & 
                \multicolumn{9}{c}{\textbf{Cognition}}\\ 
                \cmidrule(lr){3-8} \cmidrule(lr){9-17}
                
                & & 
                \multirow{2}{*}{Time} & \multirow{2}{*}{Key} & \multirow{2}{*}{ASR$\downarrow$} & \multirow{2}{*}{Visual} & \multirow{2}{*}{Camera} & 
                \multirow{2}{*}{\textbf{Avg.}} & 
                \multicolumn{3}{c}{Audio} & \multicolumn{3}{c}{Visual} & \multicolumn{2}{c}{Audio-Visual} &
                \multirow{2}{*}{\textbf{Avg.}} \\
                \cmidrule(lr){9-11} \cmidrule(lr){12-14} \cmidrule(lr){15-16} 
                
                & & & & & & & & 
                Count & Rec. & \makecell{Info.\\Ext.} & \makecell{Attr.\&\\States} & Event & Camera & Emot. & \makecell{Caus.\\Rsn.} & \\
                \midrule
                
                Gemini-3-Pro & \textbf{63.40} & \textbf{79.71} & \textbf{83.17} & \textbf{0.1631} & \underline{7.99} & \underline{34.81} & \textbf{57.87} & \textbf{48.29} & \textbf{76.82} & \textbf{46.15} & \textbf{84.38} & \textbf{81.2} & 41.92 & \textbf{88.95} & \textbf{83.67} & \textbf{68.92} \\
                
                Qwen3-Omni-30B-A3B & 45.23 & 32.01 & 45.29 & 0.5114 & \textbf{9.98} & 34.62 & 34.15 & 30.73 & 57.51 & 20.98 & 80.63 & 64.66 & \underline{50.38} & 74.21 & 71.43 & 56.32 \\
                
                Logics-Parsing-Omni(Ours) & \underline{61.12} & \underline{60.18} & \underline{64.85} & \underline{0.1698} & 6.82 & \textbf{65.58} & \underline{56.09} & \underline{39.51} & \underline{68.67} & \underline{43.36} & \underline{81.88} & \underline{68.42} & \textbf{68.65} & \underline{84.74} & \underline{73.98} & \underline{66.15} \\
                \bottomrule
            \end{tabular}
            
            \begin{tablenotes}
                \fontsize{16}{16}\selectfont 
                \item[] \textit{Note:} Bold text indicates the best result, and underlined text indicates the second-best result.
            \end{tablenotes}
        \end{threeparttable}
    }
\end{table*}
\textit{Result Analysis.} As indicated by Table~\ref{tab:logics_natural_videos_combined}, Logics-Parsing-Omni demonstrates highly competitive capabilities across both perception and cognition dimensions, achieving an Overall score of 61.12. This closely rivals the leading closed-source model Gemini-3-Pro (63.4) and significantly surpasses the Qwen3-Omni-30B-A3B baseline (45.23). 

In the \textbf{Perception} dimension, the model exhibits overwhelming superiority in \textit{Camera Motion Analysis}, attaining a joint terminology accuracy of 65.58\%, which outperforms Gemini-3-Pro (+30.77\%) and Qwen3-Omni-30B-A3B (+30.96\%) by massive margins. For fine-grained \textit{Audio-centric parsing} (e.g., identity recognition, ASR) and \textit{Temporal Grounding}, it consistently secures robust second-best results, demonstrating a solid foundation in low-level feature capture.

In the \textbf{Cognition} dimension, Logics-Parsing-Omni demonstrates highly competitive performance across diverse cross-modal reasoning tasks. Within \textit{Audio Reasoning}, it achieves a robust 39.51\% in acoustic event counting. For \textit{Visual Semantics \& Dynamics}, the model proves its exceptional capacity for macro-level semantic alignment by achieving a top-tier camera captioning accuracy of 68.65\% (+18.27\% over the strongest baseline). 

\begin{figure}[htbp]
    \centering
    \includegraphics[width=1.0\textwidth]{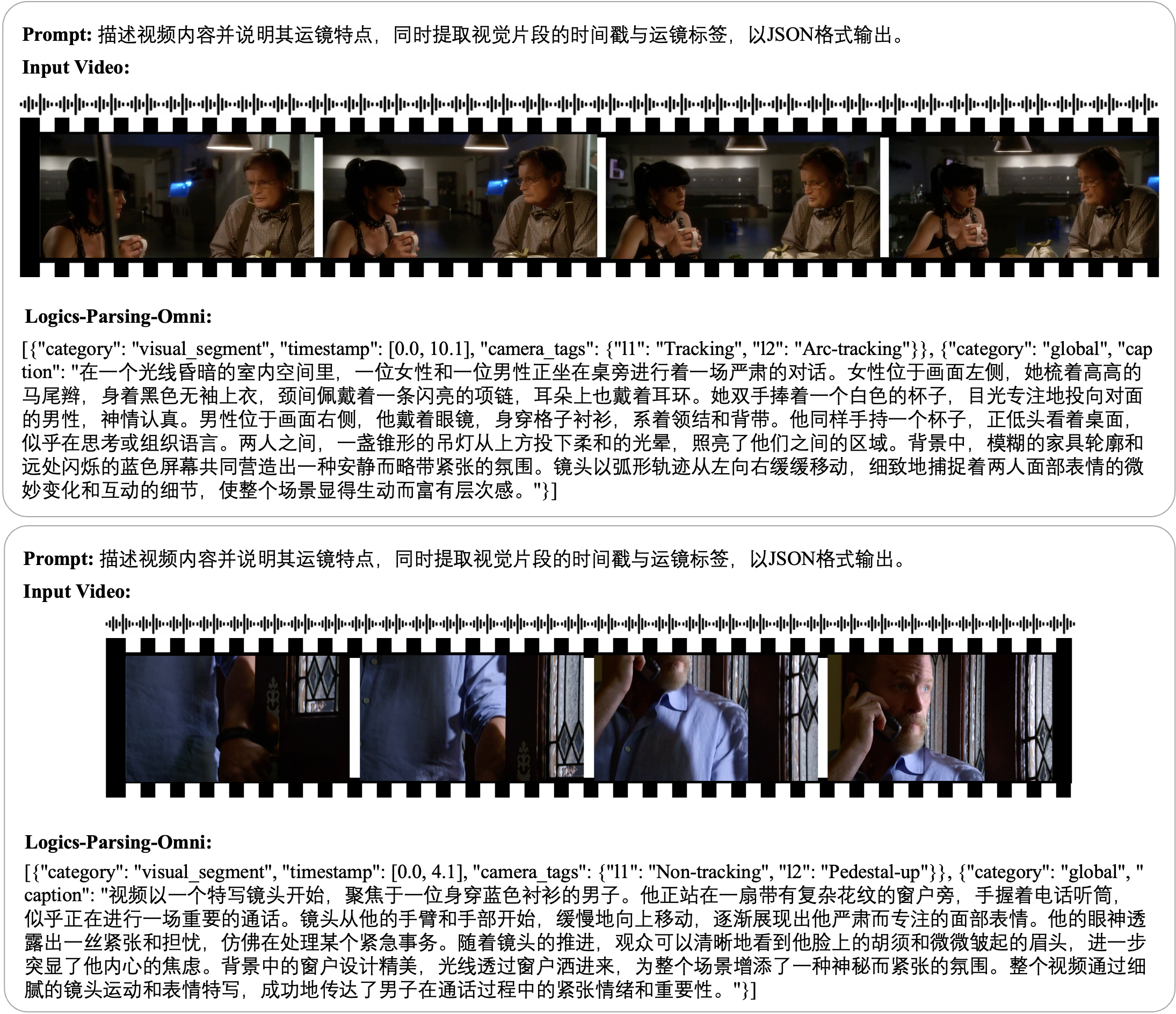}
    \caption{Qualitative examples illustrating the comprehensive camera-aware video parsing capability of Logics-Parsing-Omni.}
    \label{fig:camera_video_parsing_showcase}
\end{figure}

Qualitatively (Figures~\ref{fig:video_showcase} and Figure~\ref{fig:camera_video_parsing_showcase}), the model generates structured JSON outputs that seamlessly synchronize auditory events and hierarchical camera tags with visual sequences via high-precision timestamps. Simultaneously, it produces rich multimodal narratives that effectively weave physical spatial dynamics and the complex interplay between sound and action into cohesive global descriptions. Comprehensive details regarding the extended results are provided in Appendix~\ref{sec:appendix_videos}.

\paragraph{Text-Rich Videos.}
To comprehensively evaluate the fine-grained parsing and structured knowledge extraction capabilities of multimodal models in long-form, highly dynamic audio-visual scenarios, 
we introduce the Text-Rich Video Module of OmniParsingBench. 
Extending the framework from Section~\ref{sec:method_dataset_video}, this benchmark derives its high-fidelity ground truths by performing meticulous manual verification and refinement on the initially synthesized raw data. This process results in 259 high-quality evaluation pairs designed for parsing and structured captioning. To address both fine-grained parsing and deep understanding, we introduce a three-dimensional framework: the perception and cognition dimensions evaluate parsing performance, while the in-depth cognition dimension assesses structured captioning.

\begin{itemize}[topsep=0pt, partopsep=0pt, itemsep=2pt, parsep=0pt, leftmargin=*]
\item \textbf{Perception:} Evaluates the model's ability to accurately localize and recognize bottom-level multimodal signals, comprising three core metrics: 
    (1) \textit{Time-F1 Score}: Assesses the precision of temporal boundaries for OCR segments. We compute the temporal Intersection over Union (tIoU), and report the average F1 Score across 10 thresholds: $\{0.5, 0.55, \dots, 0.95\}$.
    (2) \textit{Visual-OCR Score}: All segment-level OCR predictions are chronologically concatenated. The score is calculated as one minus the normalized text edit distance.
    (3) \textit{Audio-ASR WER}: All segment-level ASR transcripts are concatenated to compute the Word Error Rate (WER).

\item \textbf{Cognition:} Examines semantic understanding and reasoning over the global audio-visual stream. We adopt a CaptionQA format where initial 5 challenging QAs per video are generated by Gemini-3-Pro, then refined and verified through manual annotation, conditioned on the ground truth global audio-visual description and the raw video. The model's predicted global description is then fed alongside the question into Qwen3-235B-A3B to determine answer correctness and compute the accuracy metric. The QAs span 7 fine-grained abilities across three modalities: object recognition, OCR, attributes\&states recognition, and events recognition for visual understanding; information extraction for audio understanding; along with information synthesis and causal reasoning for audio-visual understanding.

\item \textbf{Cognition (In-Depth):} Evaluates the generation of long-form, high-density structured course reports. Paralleling the standard Cognition pipeline, it utilizes ground truth in-depth structured captions to prompt Gemini-3-Pro to generate an initial set of 9 QA pairs per video, which are subsequently reviewed and refined by human annotators. Qwen3-235B-A3B then answers these QAs based on the evaluated model's predicted caption to compute an accuracy score. The QAs cover 3 categories with 9 fine-grained abilities: (1) core summary (teaching objectives, domain tags, key conclusions); (2) structured outline (structural logic, reasoning depth, and key element extraction); and (3) deep content mining (concept definitions and rigorous detail verification, requiring precise extraction of experimental variables, mathematical symbols, code parameters, or chart data).

\end{itemize}

\begin{table*}[htbp]
    \centering
    \caption{Evaluation results on the Text-Rich Video Module of OmniParsingBench}
    \label{tab:course_parsing_results}
    \renewcommand{\arraystretch}{1.3} 
    \resizebox{\textwidth}{!}{
        \begin{threeparttable}
            \begin{tabular}{l c cccc ccccccccc}
                \toprule
                \multirow{3}{*}{\textbf{Model}} & 
                \multirow{3}{*}{\textbf{Overall}} & 
                \multicolumn{4}{c}{\textbf{Perception}} & 
                \multicolumn{9}{c}{\textbf{Cognition}} \\ 
                \cmidrule(lr){3-6} \cmidrule(lr){7-15}
                & & 
                \multirow{2}{*}{Time} & \multirow{2}{*}{Visual} & \multirow{2}{*}{Audio$\downarrow$} & 
                \multirow{2}{*}{\textbf{Avg.}} & 
                Audio & \multicolumn{4}{c}{Visual} & \multicolumn{2}{c}{Audio-Visual} & 
                \multirow{2}{*}{\makecell{In-Depth\\Str. Cap.}} & 
                \multirow{2}{*}{\textbf{Avg.}} \\
                \cmidrule(lr){7-7} \cmidrule(lr){8-11} \cmidrule(lr){12-13}
                & & & & & & 
                \makecell{Info. Ext.} & 
                Object & OCR & 
                \makecell{Attr.\&States} & 
                Event & 
                \makecell{Info. Syn.} & 
                \makecell{Caus. Rsn.} & 
                & \\
                \midrule
                Gemini-3-Pro & \underline{64.37} & \underline{24.29} & \textbf{57.89} & \textbf{0.0655} & \textbf{58.54} & \underline{61.83} & \underline{23.64} & \underline{37.09} & \underline{18.67} & \underline{43.85} & \underline{63.38} & \textbf{85.12} & \textbf{92.75} & \underline{70.20} \\
                Qwen3-Omni-30B-A3B & 26.86 & 4.83 & 25.84 & 1.0242 & 10.22 & 6.22 & 1.82 & 5.63 & 2.67 & 6.92 & 8.92 & 9.30 & 81.08 & 43.50 \\
                Logics-Parsing-Omni(Ours) & \textbf{69.12} & \textbf{32.73} & \underline{57.83} & \underline{0.1839} & \underline{57.39} & \textbf{77.59} & \textbf{57.27} & \textbf{74.18} & \textbf{54.67} & \textbf{69.23} & \textbf{82.16} & \underline{84.65} & \underline{90.30} & \textbf{80.85} \\
                \bottomrule
            \end{tabular}
            
            \begin{tablenotes}
                \fontsize{16}{16}\selectfont 
                \item[] \textit{Note:} Bold text indicates the best result, and underlined text indicates the second-best result.
            \end{tablenotes}
        \end{threeparttable}
    }
\end{table*}

\begin{table*}[htbp]
    \centering
    \caption{Detailed evaluation results for the Cognition (In-Depth Str. Cap.) dimension on the Text-Rich Video Module of OmniParsingBench}
    \label{tab:course_in_depth_caption_results}
    \resizebox{\textwidth}{!}{ 
    \renewcommand{\arraystretch}{1.3} 
    \begin{tabular}{l c ccc ccc}
        \toprule
        \multirow{2}{*}{\textbf{Model}} & 
        \multirow{2}{*}{\textbf{Overall}} & 
        \multicolumn{3}{c}{\textbf{Category}} & 
        \multicolumn{3}{c}{\textbf{Video Duration}} \\
        \cmidrule(lr){3-5} \cmidrule(lr){6-8} 
        & & 
        \makecell{Core Summary} & 
        \makecell{Structured Outline} & 
        \makecell{Deep Content Mining} & 
        \makecell{Short ($<5$ min.)} & 
        \makecell{Medium (5--15 min.)} & 
        \makecell{Long (15--30 min.)} \\
        \midrule
        Gemini-3-Pro         & \textbf{92.75} & \textbf{99.23} & \textbf{93.24} & \textbf{90.64} & \underline{92.87} & \textbf{93.42} & \textbf{91.59} \\
        Qwen3-Omni-30B-A3B          & 81.08 & 94.98 & 82.43 & 76.25 & 89.99 & 82.51 & 68.57 \\
        Logics-Parsing-Omni(Ours) & \underline{90.30} & \underline{97.30} & \underline{91.51} & \underline{87.36} & \textbf{93.69} & \underline{91.56} & \underline{84.44} \\
        \bottomrule
    \end{tabular}
    } 
    \begin{tablenotes}
        \fontsize{9}{9}\selectfont 
        \item[] \textit{Note:} Bold text indicates the best result, and underlined text indicates the second-best result.
    \end{tablenotes}
\end{table*}


\textit{Result Analysis.}
As shown in Table~\ref{tab:course_parsing_results}, the overall score is aggregated from Perception and Cognition dimensions. The exact calculation rules are detailed in Appendix~\ref{sec:appendix_text_rich_videos}. The results demonstrate that our proposed Logics-Parsing-Omni achieves the highest Overall score of 69.12\%, substantially outperforming Qwen3-Omni-30B-A3B and surpassing the proprietary Gemini-3-Pro. Notably, Logics-Parsing-Omni achieves the highest average cognition score, substantially outperforming Gemini-3-Pro, while remaining competitive in fine-grained perception tasks. This highlights our model's distinct advantage in precise temporal localization and bottom-level signal recognition within highly dynamic audio-visual streams. Table~\ref{tab:course_in_depth_caption_results} reports results on the Cognition (In-Depth Str. Cap.) dimension, where the Overall score is the absolute accuracy over all QAs. Gemini-3-Pro is slightly better on overall score and long videos, while Logics-Parsing-Omni performs better on short videos. Overall, our model demonstrates robust dense knowledge extraction and long-form structured captioning, narrowing the gap between open-source models and commercial systems in educational scenarios.

To further illustrate these capabilities qualitatively, we present two showcase examples. Figure~\ref{fig:course_video_parsing_showcase} visualizes a course parsing scenario, demonstrating the model's ability to precisely localize temporal OCR segments and accurately recognize dense audio and visual-textual elements. Additionally, Figure~\ref{fig:course_video_indepth_caption_showcase} highlights its proficiency in in-depth cognition, where Logics-Parsing-Omni successfully generates highly structured, contextually rich long-form captions that encapsulate complex educational concepts.

\subsection{Performance on Public Benchmarks}
\subsubsection{OmniDocBench v1.5} 
Apart from the previously proposed LogicsDocBench, we also conduct experiments on the commonly used open-sourced OmniDocBench-v1.5~\cite{niu2025mineru2} benchmark for more fair and intuitive comparisons. Compared to its previous version (v1.0), OmniDocBench-v1.5 incorporates 374 additional document pages, resulting in an expanded test set of 1,355 pages with a more balanced distribution between Chinese and English content. Furthermore, the benchmark introduces updated evaluation protocols, where Character Detection Matching (CDM)~\cite{wang2025image} methods are employed for math formulas recognition evaluation, as well as the overall scores are calculated by weighted combination of the text edit distance, formula CDM score and the table Tree Edit Distance Similarity (TEDS) score.

\begin{table*}[h]
\centering
\caption{Comparisons with State-of-the-art methods on OmniDocBench-v1.5.}
\label{tab:omnidocbench_results}
\resizebox{\textwidth}{!}{%
    \begin{tabular}{llcccccc}
    \toprule 
    Model Type & Methods & Overall $\uparrow$ & Text Edit $\downarrow$ & Formula CDM $\uparrow$ & Table TEDS $\uparrow$ & Table TEDS-S $\uparrow$ & ReadOrder $\downarrow$ \\
    \midrule
    \multirow{3}{*}{Pipeline Tools} & Marker-1.8.2~\footnotemark[4] & 71.3 & 0.206 & 76.66 & 57.88 & 71.17 & 0.250 \\
    & MinerU2-Pipe~\cite{wang2024mineru} & 75.51 & 0.209 & 76.55 & 70.90 & 79.11 & 0.225 \\
    & PP-StructureV3~\cite{cui2025paddleocr30technicalreport} & 86.73 & 0.073 & 85.79 & 81.68 & 89.48 & 0.073 \\
    \midrule
    \multirow{11}{*}{General MLLMs} & GPT-4o~\footnotemark[10] & 75.02 & 0.217 & 79.70 & 67.07 & 76.09 & 0.148 \\
     & InternVL3~\cite{zhu2025internvl3} & 80.33    & 0.131 & 83.42 & 70.64 & 77.74 & 0.113 \\
     & InternVL3.5~\cite{wang2025internvl3} & 82.67 & 0.142 & 87.23 & 75.00 & 81.28 & 0.125 \\
     & GPT-5.2~\footnotemark[9] & 86.23 & 0.115 & 88.6 & 81.60 & 86.42 & 0.099 \\
     & Qwen2.5-VL-72B~\cite{bai2025qwen2} & 87.02 & 0.094 & 88.27 & 82.15 & 86.22 & 0.102 \\
     & Gemini-2.5-Pro~\cite{comanici2025gemini} & 88.03 & 0.075 & 85.82 & 85.71 & 90.29 & 0.097 \\
     & Gemini-3-Pro~\footnotemark[13] & 88.36 & 0.062 & 83.1 & 88.19 & 93.52 & 0.072 \\
     & Qwen3-VL-30B-A3B~\cite{Qwen3-VL} & 83.61 & 0.056 & 79.4 & 77.04 & 81.94 & 0.081\\
     & Qwen3-VL-235B-A22B~\cite{Qwen3-VL} & 89.15 & 0.069 & 88.14 & 86.21 & 90.55 & 0.068\\
     & Qwen3-Omni (Base)~\cite{Qwen3-Omni} & 74.84 & 0.137 & 64.0 & 74.21 & 77.68 & 0.117 \\
     & Qwen3.5-397b-a17b~\footnotemark[16]        & 88.34 & 0.055 & 87.6 & 82.93 & 88.70 & 0.080 \\ 
     \midrule
     \multirow{16}{*}{Specialized MLLMs} & Dolphin~\cite{feng2025dolphin}  & 74.67 & 0.125 & 67.85 & 68.70 & 77.77 & 0.124 \\
     & OCRFlux~\footnotemark[7]  & 74.82 & 0.193 & 68.03 & 75.75 & 80.23 & 0.202 \\
     & Mistral OCR~\footnotemark[10] & 78.83 & 0.164 & 82.84 & 70.03 & 78.04 & 0.144 \\
     & POINTS-Reader~\cite{liu2025points} & 80.98 & 0.134 & 79.20 & 77.13 & 81.66 & 0.145 \\
     & Dolphin-1.5~\cite{feng2025dolphin}  & 83.21 & 0.092 & 80.78 & 78.06 & 84.10 & 0.080 \\
     & olmOCR~\cite{poznanski2025olmocr}  & 81.79 & 0.096 & 86.04 & 68.92 & 74.77 & 0.121 \\
     & MinerU2-VLM~\cite{wang2024mineru}  & 85.56 & 0.078 & 80.95 & 83.54 & 87.66 & 0.086 \\
     & Nanonets-OCR-s~\cite{Nanonets-OCR-S} & 85.59 & 0.093 & 85.90 & 80.14 & 85.57 & 0.108 \\
     & MonkeyOCR-pro-1.2B~\cite{li2025monkeyocrdocumentparsingstructurerecognitionrelation}  & 86.96 & 0.084 & 85.02 & 84.24 & 89.02 & 0.130 \\
     & Deepseek-OCR~\cite{wei2025deepseek} & 87.01 & 0.073 & 83.37 & 84.97 & 88.80 & 0.086 \\
     & MonkeyOCR-3B~\cite{li2025monkeyocrdocumentparsingstructurerecognitionrelation}  & 87.13 & 0.075 & 87.45 & 81.39 & 85.92 & 0.129 \\
     & dots.ocr~\footnotemark[6]  & 88.41 & 0.048 & 83.22 & 86.78 & 90.62 & 0.053 \\
     & OCRVerse~\footnotemark[11] & 88.56 & 0.058 & 86.91 & 84.55 & 88.45 & 0.071 \\
     & MonkeyOCR-pro-3B~\cite{li2025monkeyocrdocumentparsingstructurerecognitionrelation} & 88.85 & 0.075 & 87.25 & 86.78 & 90.63 & 0.128 \\
     & MinerU2.5~\cite{niu2025mineru2} & 90.67 & \underline{0.047} & 88.46 & 88.22 & 92.38 & \underline{0.044} \\
     & PaddleOCR-VL~\cite{cui2025paddleocrvlboostingmultilingualdocument} & \textbf{92.86} & \textbf{0.035} & \underline{91.22} & \underline{90.89} & \textbf{94.76} & \textbf{0.043} \\

    \midrule
    General MLLMs & \textbf{Logics-Parsing-Omni (Ours)} & \underline{92.54} & \underline{0.047} & \textbf{91.3} & \textbf{91.02} & \underline{94.31} & 0.056 \\
    \bottomrule
    \end{tabular}%
}

\begin{minipage}{\textwidth}
\vspace{1ex} 
\textit{Note:} Bold text indicates the best result, and underlined text indicates the second-best result. 
\end{minipage}
\end{table*}

\textit{Result Analysis.}
The experimental results of our model are presented in Table.~\ref{tab:omnidocbench_results}. Our Logics-Parsing-Omni model performs only slightly inferior to the light-weight specialized VLM PaddleOCR-VL~\cite{cui2025paddleocrvlboostingmultilingualdocument} on the overall scores, while outperforming all other models with general or specialized document-parsing capabilities. As a two-stage pipeline which separates layout analysis from element recognition, PaddleOCR-VL can perform the fine-grained optimization of the element recognition stage with a smaller and more efficient network. In contrast, Logics-Parsing-Omni integrates both layout analysis and element recognition into a unified single-stage framework. Moreover, it is designed as a general-purpose model capable of handling parsing tasks across diverse modalities, including text, images, video, and audio, which imposes significantly larger challenges for the whole model design and optimization. Despite this, Logics-Parsing-Omni still achieves the best scores on formula and table recognition, and only slightly inferior on text edit distance and table TEDS-S metric, which provides strong evidence on the effectiveness of our omni model architecture, as well as its data construction and training strategy.

\subsection{Advanced Multi-image Analysis}
\subsubsection{Natural Image}

To comprehensively evaluate the model's performance in image difference recognition and localization, we curated a test set of 500 samples from the dataset described in Section~\ref{sec:nature_image_dataset}. This test set ensures diversity in scene coverage and a balanced distribution of difference operation types. To achieve fine-grained evaluation, we assess the model across two dimensions:
\begin{itemize}[topsep=0pt, partopsep=0pt, itemsep=2pt, parsep=0pt, leftmargin=*]
\item \textbf{Difference Perception:} Precision and Recall are calculated based on the Intersection over Union (IoU) between predicted and ground truth bounding boxes, subject to a predefined threshold.
\item \textbf{Difference Cognition:} To quantify discrepancies between predicted and reference descriptions more precisely, we employ an \textit{LLM-as-a-Judge} mechanism. Specifically, both predicted and reference captions are parsed into lists of difference points. Precision and Recall are then computed by matching these lists. Furthermore, predicted difference points that cannot be matched to reference points due to semantic variations are categorized as \textit{irrelevant items}, enabling the calculation of an \textit{Irrelevance Rate} to comprehensively reflect generation quality.\end{itemize}

\begin{figure}[htbp]
    \centering
    \includegraphics[width=1.0\textwidth]{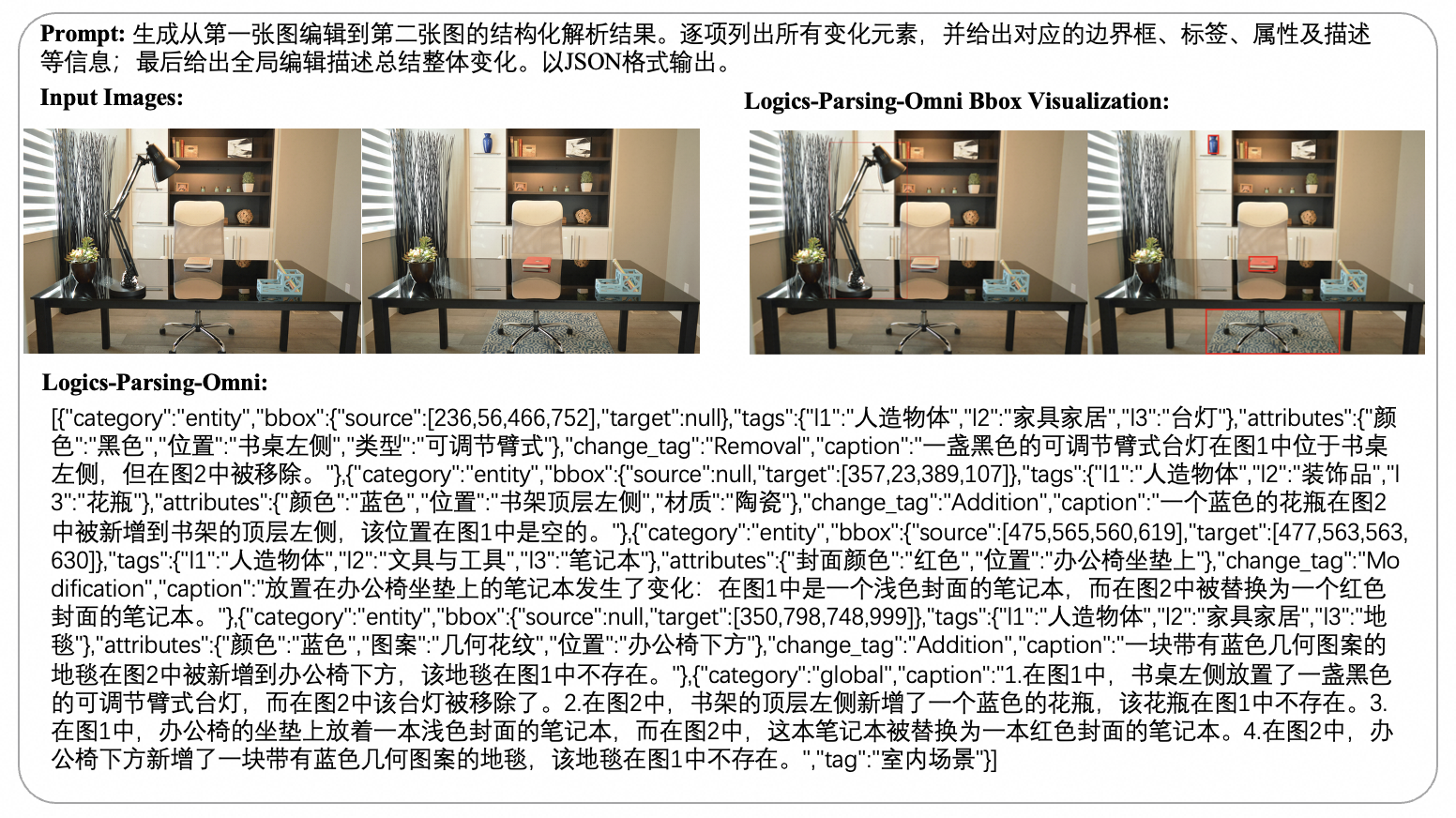}
    \caption{Qualitative examples illustrating the natural image difference parsing capability of Logics-Parsing-Omni. }
    \label{fig:diff_parsing_view_1}
\end{figure}

\textit{Result Analysis.}
Experimental results presented in Table~\ref{tab:idcg} indicate that our model significantly outperforms Qwen3-Omni-30B-A3B  across both evaluation dimensions, particularly in the difference localization task. Moreover, our model's precision surpasses that of current state-of-the-art (SOTA) commercial closed-source models, such as Gemini-3-Pro. As shown in Figure ~\ref{fig:diff_parsing_view_1}, our model successfully detects inter-image differences and accurately localizes these regions using bounding boxes, highlighting its robust performance in image difference tasks.
\begin{table*}[htbp]
\centering
\scriptsize 
\renewcommand{\arraystretch}{1.2} 
\caption{Evaluation results on the Natural-Image Difference Module of OmniParsingBench}
\label{tab:idcg}
\begin{tabular*}{\textwidth}{@{\extracolsep{\fill}}lccccc}
\toprule
\multirow{2}{*}{\textbf{Model}}  & \multicolumn{2}{c}{\textbf{Perception}} & \multicolumn{3}{c}{\textbf{Cognition}} \\
\cmidrule(lr){2-3} \cmidrule(lr){4-6}

& Precision ($\uparrow$) & Recall ($\uparrow$) & Precision ($\uparrow$) & Irre. ($\downarrow$) & Recall ($\uparrow$) \\
\midrule
Gemini-3-Pro & 0.417 & 0.349 & 0.573 & 0.331 & 0.460  \\
GPT-5.2 & 0.123 & 0.169 & \textbf{0.671} & 0.414 & 0.490  \\
Qwen3.5-397B-A17B & 0.383 & 0.328 & 0.621 & 0.376 & 0.461 \\
Qwen3-VL-235B-A22B & 0.356 & 0.293 & 0.653 & 0.339 & \textbf{0.518} \\
Qwen3-VL-30B-A3B & 0.242 & 0.315 & 0.496 & 0.402 & 0.365  \\
Qwen3-Omni-30B-A3B & 0.213 & 0.327 & 0.435 & 0.365 & 0.317  \\
\midrule
Logics-Parsing-Omni(Ours) & \textbf{0.689} & \textbf{0.533} & 0.609 & \textbf{0.180} & 0.489  \\
\bottomrule
\end{tabular*}
\end{table*}

\subsubsection{Graphics}  
To evaluate a model’s ability to perceive fine-grained visual differences and articulate precise editing steps, 
we introduce the Geo-Image Difference Module of OmniParsingBench. This benchmark assesses the model's performance on a dual-objective task: generating structured Plotting data (coordinates and relations) and Global natural language instructions.

The dataset comprises 1000 human-calibrated image pairs. We establish a hierarchical evaluation framework, as detailed in Table~\ref{tab:geo_edit_results}, comprising two primary dimensions: Factual Perception and Semantic Cognition.

\begin{itemize}[topsep=0pt, partopsep=0pt, itemsep=3pt, parsep=0pt, leftmargin=*]
\item \textbf{Perception.} Focuses on visual grounding precision. This includes \textbf{Difference Detection} (numerical precision of coordinates and radii) and \textbf{Difference Identification} (existence of primitives and recognition of geometric/quantitative relations).

\item \textbf{Cognition.} Assesses the reasoning quality of instructions. It consists of \textbf{Accuracy} (correct entity naming and geometric verbs) and \textbf{Plausibility} (logical completeness, sequence dependency, and absence of hallucinations).
\end{itemize}

\begin{table*}[htbp]
\centering
\scriptsize 
\renewcommand{\arraystretch}{1.25} 
\setlength{\tabcolsep}{3.5pt} 
\caption{Evaluation results on the Geo-Image Difference Module of OmniParsingBench.}
\label{tab:geo_edit_results}
\resizebox{\textwidth}{!}{%
\begin{tabular}{lcccccccccccc}
\toprule
\multirow{4}{*}{\textbf{Model}} & \multirow{4}{*}{\textbf{Overall}} & \multicolumn{7}{c}{\textbf{Perception}} & \multicolumn{4}{c}{\textbf{Cognition}} \\
\cmidrule(lr){3-9} \cmidrule(lr){10-13}

 & & \multicolumn{2}{c}{Diff Detection} & \multicolumn{5}{c}{Diff Identification} & \multicolumn{2}{c}{Accuracy} & \multicolumn{2}{c}{Plausibility} \\
\cmidrule(lr){3-4} \cmidrule(lr){5-9} \cmidrule(lr){10-11} \cmidrule(lr){12-13}

 & & \multicolumn{2}{c}{Positioning} & \multicolumn{3}{c}{Existence} & \multicolumn{2}{c}{Relation} & \multirow{2}{*}{Object} & \multirow{2}{*}{Operation} & \multirow{2}{*}{Logic} & \multirow{2}{*}{Halluc.} \\
\cmidrule(lr){3-4} \cmidrule(lr){5-7} \cmidrule(lr){8-9}

 & & Coord. & Radius & Point & Line/Arc & Circle & Geo. & Quant. & & & & \\
\midrule

Gemini-3-Pro & 27.35 & 26.45 & 0.00 & 32.23 & 1.05 & 1.59 & 4.71 & 2.71 & 37.16 & 33.92 & 34.74 & 37.42 \\
GPT-5.2 & 36.43 & 7.46 & 0.47 & 58.55 & 9.02 & 18.98 & 6.18 & 5.76 & 50.16 & 43.74 & 46.98 & 47.94 \\
Qwen3.5-397B-A17B & 33.68 & 11.92 & 0.87 & 44.54 & 2.60 & 6.54 & 6.34 & 4.94 & 46.84 & 43.22 & 44.62 & 45.88 \\
Qwen3-VL-235B-A22B & 32.66 & 13.43 & 0.78 & 59.21 & 5.39 & 18.86 & 4.89 & 4.34 & 42.28 & 36.36 & 39.12 & 41.48 \\
Qwen3-VL-30B-A3B & 29.46 & 36.23 & 1.08 & 57.83 & 5.78 & 14.31 & 2.49 & 2.10 & 33.30 & 29.10 & 31.44 & 34.64 \\
Qwen3-Omni-30B-A3B & 31.24 & 22.91 & 5.94 & 58.98 & 4.30 & 34.57 & 4.17 & 3.60 & 36.56 & 31.70 & 34.20 & 37.52 \\
\midrule
Logics-Parsing-Omni(Ours) & \textbf{52.81} & \textbf{43.25} & \textbf{20.28} & \textbf{69.24} & \textbf{13.30} & \textbf{63.70} & \textbf{22.04} & \textbf{12.12} & \textbf{63.60} & \textbf{59.32} & \textbf{60.52} & \textbf{68.18} \\

\bottomrule
\end{tabular}%
}
\end{table*}

\begin{figure}[htbp]
    \centering
    \includegraphics[width=1.0\textwidth]{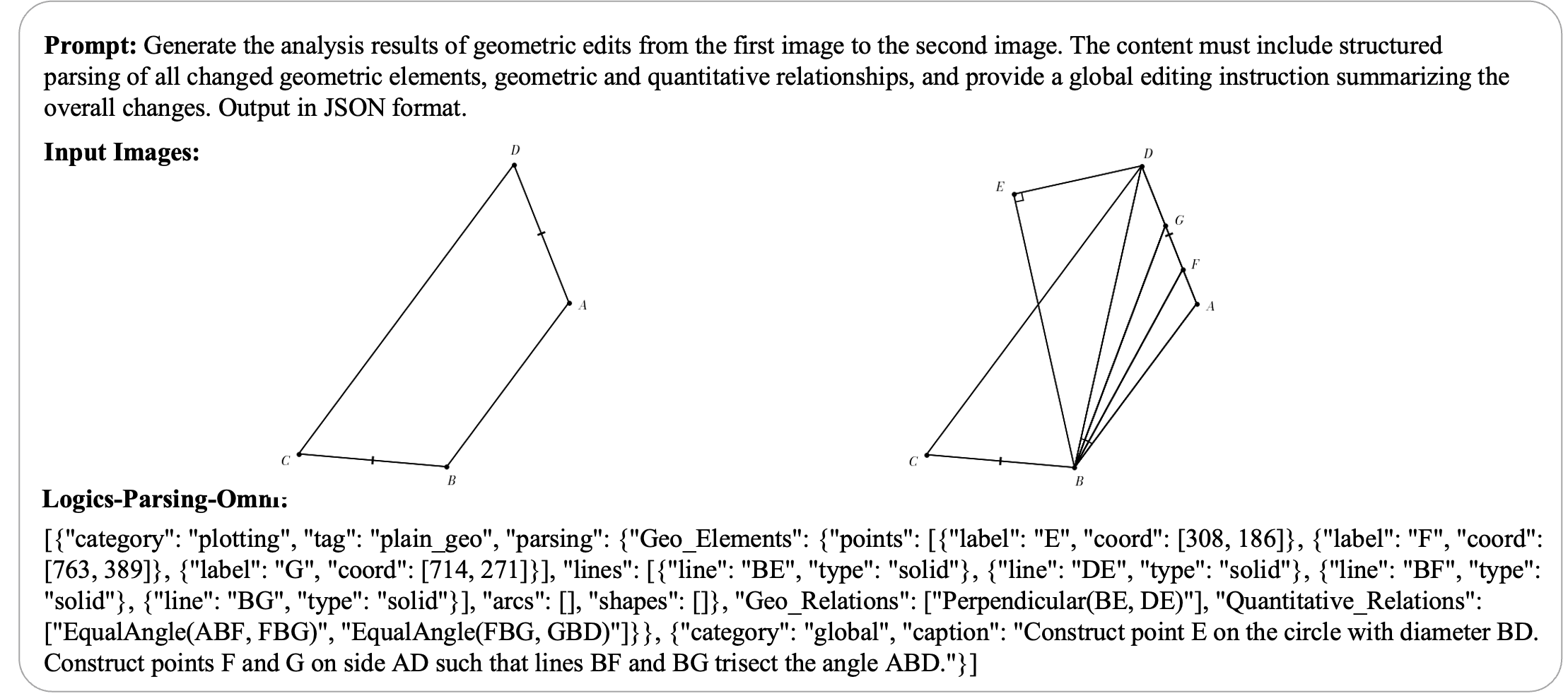}
    \caption{Qualitative examples illustrating the graphic image difference parsing capability of Logics-Parsing-Omni. }
    \label{fig:geo_edit_case}
\end{figure}

\textit{Result Analysis.}
As shown in Table~\ref{tab:geo_edit_results}, Logics-Parsing-Omni achieves an overall score of 52.81, significantly outperforming the baseline Qwen3-Omni-30B-A3B (31.24). Our model demonstrates superior performance across both perception and cognition metrics, particularly in high-precision tasks like coordinate positioning and geometric relation identification. As shown in Figure ~\ref{fig:geo_edit_case}, Our model has achieved highly precise parsing and captioning capabilities on geometric difference diagrams, accurately locating the points of difference, describing the newly added geometric/quantitative relationships, and providing precise summarizations in natural language.

\section{Discussion} 

\begin{table*}[h]
\centering
\caption{Ablation results on the Graphics Cognition.}
\label{tab:ablation_exp}
\renewcommand{\arraystretch}{1.3}
\resizebox{\textwidth}{!}{
\begin{tabular}{l c cccc cccc}
\toprule
\multicolumn{1}{c}{\multirow{2.6}{*}{\textbf{Model}}} & \multirow{2.6}{*}{\textbf{Overall}} & \multicolumn{4}{c}{\textbf{Chart}} & \multicolumn{4}{c}{\textbf{Geometry}} \\
\cmidrule(lr){3-6} \cmidrule(lr){7-10}
& & \makecell{OCR \\ \& Data} & \makecell{Visual \\ Elements} & \makecell{Logical \\ Reasoning} & \makecell{\textbf{Avg.}} & \makecell{Basic \\ Elements} & \makecell{Geometric \\ Relations} & \makecell{Quantitative \\ Relations} & \makecell{\textbf{Avg.}} \\
\midrule
Qwen3-Omni-30B-A3B   & 79.55 & 82.33 & 86.51 & 73.97 & 81.40 & 80.16 & 71.67 & 80.41 & 77.70 \\
+ Graphics (Caption Only)   & 79.70 & 81.80 & 87.91 & 68.04 & 80.10 & 84.10 & 70.65 & 75.26 & 79.30 \\
+ Graphics (Parsing \& Caption) & \textbf{91.80} & \textbf{97.17} & \textbf{95.81} & \textbf{90.87} & \textbf{95.50} & \textbf{90.16} & \textbf{82.94} & \textbf{90.72} & \textbf{88.10} \\
\bottomrule
\end{tabular}
}
\end{table*}

To deeply validate the synergistic effectiveness between fine-grained structural parsing (Perception) and high-level semantic reasoning (Cognition), we conducted comprehensive ablation studies. The ablation models in this section are fine-tuned exclusively on our curated information-centric graphics corpus. We initialized our experiments from the zero-shot baseline model, Qwen3-Omni-30B-A3B. To evaluate the gains brought by structured data, we trained two variants. The first variant, \textbf{+ Graphics (Caption Only)}, was fine-tuned exclusively on global semantic descriptions, completely bypassing the extraction of fine-grained perception data. The second variant, \textbf{+ Graphics (Parsing \& Caption)}, adopted our complete structural paradigm, which deeply integrates fine-grained perception data (eg, HTML reverse-rendering for charts, explicit topologies for geometry) with rich semantic descriptions.

Table~\ref{tab:ablation_exp} details the performance across cognition dimensions. We observe that fine-tuning with captions alone yields marginal gains, resulting in an Overall score of 79.70. In contrast, introducing complete structural parsing triggers a substantial performance leap, driving the Overall accuracy to 91.80.

An analysis of the fine-grained dimensional metrics reveals that, in Chart Cognition tasks, relying solely on natural language descriptions actively harms the model's Logical Reasoning capability, causing it to drop from 73.97 to 68.04. However, when the model is forced to learn fine-grained structural information (the Parsing \& Caption variant), this metric surges to 90.87 (an absolute increase of 22.83). This proves that explicit structured anchors are crucial in helping the model perform reliable logical reasoning over implicit quantitative trends. 

In the geometric domain, pure caption training similarly leads to a degradation in the Quantitative Relations metric (dropping from 80.41 to 75.26). This decline suggests that feeding the model free-form spatial descriptions without foundational anchors actually disrupts its perception of metric constraints. Conversely, introducing explicit structural parsing (which strictly defines primitive topologies and coordinates) not only rectifies this confusion—elevating Geometric Relations to 82.94, but also substantially boosts the extraction precision of Quantitative Relations to 90.72.

The aforementioned ablation results demonstrate that translating abstract visual signals into enumerable, structurally faithful fine-grained perception data serves as an indispensable foundation for achieving reliable, high-level semantic cognition.

\section{Conclusions}
In this paper, we presented Logics-Parsing-Omni, a unified multimodal framework that effectively bridges the chasm between pixel-level perception and semantic cognition. By implementing the Omni Parsing paradigm—supported by a meticulously curated standardized corpus and a two-stage progressive training strategy—we have successfully transformed unstructured heterogeneous data (spanning documents, images, and audio-visual streams) into locatable, enumerable, and traceable knowledge. Empirical results on OmniParsingBench confirm that our model not only achieves state-of-the-art performance across diverse modalities but also validates the synergy between fine-grained structural fidelity and deep logical reasoning. Moving forward, we aim to further advance this frontier by optimizing computational efficiency and developing robust cross-modal continual learning mechanisms, paving the way for more scalable and adaptive knowledge-intensive AI systems.

\section{Contribution}
The contributors’ names are sorted in alphabetical order of the last name.

\textbf{Core Contributors} 

Xin An, Jingyi Cai, Xiangyang Chen, Huayao Liu, Peiting Liu, Peng Wang, Bei Yang, Xiuwen Zhu

\textbf{Contributors} 

Yongfan Chen, Yan Gao, Yuan Gao, Baoyu Hou, Guangzheng Hu, Shuzhao Li, Weixu Qiao, Weidong Ren, Yanan Wang, Boyu Yang, Fan Yang, Jiangtao Zhang, Lixin Zhang

\textbf{Project Leads} 

Lin Qu, Hu Wei, Xiaoxiao Xu, Bing Zhao

\bibliography{references}
\bibliographystyle{technical_report}
\appendix 
\section{Appendix}

\subsection{Detailed Evaluation on the Natural Image Module of OmniParsingBench}\label{subsec:appendix_logics_image_parsing}

\noindent\textbf{Perception Analysis.}
The perception metric computes the instance-level parsing score. An instance (entity, knowledge-aware entity, or text) is considered correct \textit{only if}: (1) its predicted bounding box achieves the highest Intersection-over-Union (IoU) with a ground-truth box (minimum IoU $\ge$ 0.5), \textit{and} (2) all associated fields pass verification. The verification criteria are as follows: LLM-verified semantic consistency for standard entities; LLM-verified knowledge accuracy for knowledge-aware entities (note that although the parsing output lacks explicit knowledge tags, models may interpret knowledge boundaries differently, necessitating LLM-based judgment); and exact string matching for OCR (string similarity $> 0.7$ based on normalized edit distance) alongside language identification for text. The score is calculated as: $\text{Score} = \frac{\text{correctly parsed instances}}{\text{total ground-truth instances}}$. As shown in Table~\ref{tab:image_parsing_perception_detail_results}, Logics-Parsing-Omni demonstrates significant progress in fine-grained structural parsing, achieving balanced improvements in both localization capability and semantic fidelity, which enables more precise scene decomposition.

However, knowledge-aware entity parsing remains a key challenge: despite strong localization recall (92.25\%), semantic accuracy drops to 48.99\% due to difficulties in precisely referencing verified knowledge identifiers. This gap highlights the need for targeted strategies—such as knowledge-grounded data synthesis, contrastive identifier training, external knowledge retrieval integration—to strengthen factual grounding without compromising structural parsing integrity.

\begin{table}[htbp]
\centering
\caption{Detailed Evaluation Results on the Natural Image module of OmniParsingBench - Perception}
\label{tab:image_parsing_perception_detail_results}
\resizebox{\textwidth}{!}{
\begin{tabular}{l ccc ccc cccc c}
\toprule
\multicolumn{1}{c}{\multirow{2.6}{*}{\textbf{Model}}}
& \multicolumn{3}{c}{\textbf{Entity}} 
& \multicolumn{3}{c}{\textbf{Knowledge-Aware Entity}} 
& \multicolumn{4}{c}{\textbf{Text}}
& \multicolumn{1}{c}{\multirow{2.6}{*}{\textbf{Average}}}\\
\cmidrule(lr){2-4} \cmidrule(lr){5-7} \cmidrule(lr){8-11} 
& Loc. Recall & Sem. Acc.$^*$ & \textbf{Score} 
& Loc. Recall & Sem. Acc.$^*$ & \textbf{Score} 
& Loc. Recall & OCR Acc.$^*$ & Lang. Acc.$^*$ & \textbf{Score}  \\
\midrule
Gemini 3 Pro          & \textbf{75.90} & 97.35 & \textbf{73.89} & 88.01 & \textbf{59.44} & 52.32 & 59.09 & 74.49 & 88.46 & 41.67 & 55.96 \\
GPT 5.2$^\dagger$       & 58.30 & 93.08 & 54.26 & - & - & - & 47.52 & 49.25 & 86.57 & 21.28 & 37.77\\
Qwen3.5-397B-A17B     & 73.54 & 97.60 & 71.77 & 89.30 & 51.80 & 46.26 & 83.10 & 74.58 & 84.70 & \textbf{52.82} & \textbf{56.95} \\
Qwen3-VL-235B-A22B-Instruct    & 68.56 & 97.41 & 66.79 & 89.30 & 58.68 & \textbf{52.41} & 76.11 & 69.30 & 85.96 & 49.50 & 56.23\\
Qwen3-VL-30B-A3B-Instruct      & 63.08 & \textbf{97.65} & 61.60 & 83.15 & 44.77 & 37.23 & \textbf{88.03} & 64.80 & 83.20 & 47.89 & 48.91\\
Qwen3-Omni-30B-A3B-Instruct (base) & 71.76 & 97.15 & 69.72 & 84.45 & 33.02 & 27.88 & 71.13 & 67.33 & \textbf{90.10} & 42.96 & 46.85\\
\midrule
\textbf{Logics Parsing Omni (ours)} & 72.83 & 97.55 & 71.05 & \textbf{92.25} & 48.99 & 45.19 & 72.54 & \textbf{74.76} & 83.50 & 45.07 & 53.77\\
\bottomrule
\multicolumn{12}{l}{$^*$ Sem. Acc., OCR Acc. and Lang. Acc. are computed conditionally on successfully localized instances. Localization Recall and final Score metrics use total instances as denominator.} \\
\multicolumn{12}{l}{$^\dagger$ Note that GPT seems to fail to identify some knowledge-aware entities due to its model strategy, knowledge metrics are ignored here.}\\
\end{tabular}
}
\end{table}

\noindent\textbf{Cognition Analysis.}
Cognition metrics evaluate global caption quality via two components: (1) \textit{Caption QA Accuracy}: Human-annotated question sets aligned with nine semantic aspects (theme, emotion, scene, etc.) are answered by an LLM using \textit{only} the model-generated caption; per-aspect accuracy is macro-averaged across images. (2) \textit{Knowledge Mention Accuracy}: Proportion of knowledge-annotated images where the caption explicitly and correctly references \textit{all} verified knowledge entities (LLM-verified). 

Table~\ref{tab:image_parsing_cognition_detail_results} presents detailed evaluation results on cognition Caption QA. The \textit{Answerable} metric quantifies caption comprehensiveness (proportion of QA pairs addressable by the caption), while \textit{Precision} reflects factual accuracy of answered questions—serving as a direct indicator of hallucination control. Logics Parsing Omni achieves leading performance across all semantic aspects in parsing-derived caption evaluation, with the highest Answerable rate and robust Precision among open-weight models, demonstrating an optimal balance between informational coverage and semantic fidelity.

A critical observation is that parsing-derived captions from most models are significantly shorter than their direct dense captions (e.g., Gemini-3-Pro: 1613 vs. 571 (string length); Qwen3-Omni-30B-A3B: 1055 vs. 404), potentially introducing length-related bias in caption quality assessment. To ensure a fair and length-agnostic comparison, we additionally evaluate all models’ direct dense captioning capability under identical QA protocols. The results confirm that Logics-Parsing-Omni maintains clear leadership: its parsing-generated caption achieves a 82.60\% overall QA score—surpassing all open-weight baselines’ dense captions. This demonstrates that knowledge-enhanced structured parsing inherently produces captions with superior semantic density and factual reliability, without relying on verbose generation. The framework effectively decouples caption quality from length, validating its robustness for high-fidelity visual cognition.

\begin{table}[htbp]
\centering
\footnotesize 
\caption{Detailed Evaluation Results on the Natural Image module of OmniParsingBench - Cognition.}
\label{tab:image_parsing_cognition_detail_results}
\resizebox{\linewidth}{!}{
    \begin{tabular}{l ccc c ccccccccc}
    \toprule
    \multirow{2}{*}{\textbf{Model}} & 
    \multicolumn{3}{c}{\textbf{Overall Metrics}} & 
    \multirow{2}{*}{\textbf{\makecell{Caption \\ Length}}} & 
    \multicolumn{9}{c}{\textbf{Detailed Evaluation Metrics}} \\
    \cmidrule(lr){2-4} \cmidrule(lr){6-14}
    & \textbf{Overall} & \textbf{Answerable} & \textbf{Precision} & & 
    \textbf{Theme} & \textbf{Emotion} & \textbf{Scene} & \textbf{Count} & \textbf{Attribute} & \textbf{Location} & \textbf{Text} & \textbf{Style} & \textbf{Composition} \\
    \midrule
    \textit{Direct Dense Caption} \\
    Gemini 3 Pro & \textbf{86.38} & \textbf{93.40} & \textbf{92.48} & 1{,}613 & \textbf{95.43} & 86.27 & \textbf{87.54} & \textbf{80.55} & \textbf{90.06} & \textbf{83.87} & \textbf{90.16} & \textbf{84.72} & \textbf{77.55} \\
    GPT 5.2 & 73.52 & 81.83 & 89.86 & \textbf{1{,}726} & 89.12 & 49.02 & 79.44 & 63.64 & 75.22 & 73.94 & 59.39 & 78.12 & 68.54 \\
    Qwen3.5-397B-A17B & 82.48 & 90.32 & 91.33 & 1{,}060 & 95.23 & 88.24 & 86.20 & 71.50 & 86.72 & 80.07 & 79.19 & 80.06 & 74.10 \\
    Qwen3-VL-235B-A22B-Instruct & 79.86 & 88.75 & 89.98 & 1{,}234 & 94.40 & 89.22 & 86.76 & 68.47 & 82.94 & 75.44 & 80.71 & 78.67 & 69.82 \\
    Qwen3-VL-30B-A3B-Instruct & 77.20 & 86.41 & 89.34 & 1{,}265 & 94.16 & \textbf{92.16} & 85.63 & 65.11 & 79.95 & 70.11 & 68.53 & 78.39 & 69.48 \\
    Qwen3-Omni-30B-A3B-Instruct & 76.75 & 85.88 & 89.37 & 1{,}055 & 94.16 & 91.18 & 85.63 & 61.43 & 80.67 & 69.86 & 70.05 & 78.67 & 66.93 \\
    \midrule
    \textit{Caption from Image parsing} \\
    Gemini 3 Pro & 74.14 & 81.15 & \textbf{91.37} & 571 & \textbf{95.20} & 85.29 & \textbf{87.25} & 60.00 & 72.68 & 72.14 & 66.84 & 74.58 & 66.19 \\
    GPT 5.2 & 66.46 & 74.12 & 89.66 & 434 & 93.35 & 50.00 & 81.64 & 50.74 & 58.66 & 67.62 & 59.90 & 75.77 & 63.93 \\
    Qwen3.5-397B-A17B & 66.41 & 74.12 & 89.59 & 445 & 92.04 & 78.43 & 82.49 & 50.61 & 62.50 & 61.68 & 59.18 & 71.47 & 59.44 \\
    Qwen3-VL-235B-A22B-Instruct & 68.26 & 74.92 & 91.11 & 556 & 93.90 & 84.31 & 84.23 & 51.48 & 65.70 & 60.32 & 63.45 & 73.13 & 61.04 \\
    Qwen3-VL-30B-A3B-Instruct & 63.75 & 71.31 & 89.41 & 473 & 93.51 & 83.17 & 83.38 & 49.04 & 60.50 & 50.70 & 48.70 & 72.40 & 57.17 \\
    Qwen3-Omni-30B-A3B-Instruct & 61.65 & 68.98 & 89.37 & 404 & 92.60 & 82.47 & 81.45 & 46.65 & 59.08 & 49.45 & 45.70 & 67.52 & 53.50 \\
    \textbf{Logics Parsing Omni (Ours)} & \textbf{82.60} & \textbf{90.65} & 91.12 & \textbf{1{,}690} & 94.40 & \textbf{92.16} & 87.01 & \textbf{71.18} & \textbf{85.65} & \textbf{79.54} & \textbf{77.04} & \textbf{83.38} & \textbf{76.01} \\
    \bottomrule
    \end{tabular}
}
\end{table}

\noindent\textbf{Limitations and Future Directions.}

\begin{figure}[htbp]
    \centering
    \includegraphics[width=1.0\textwidth]{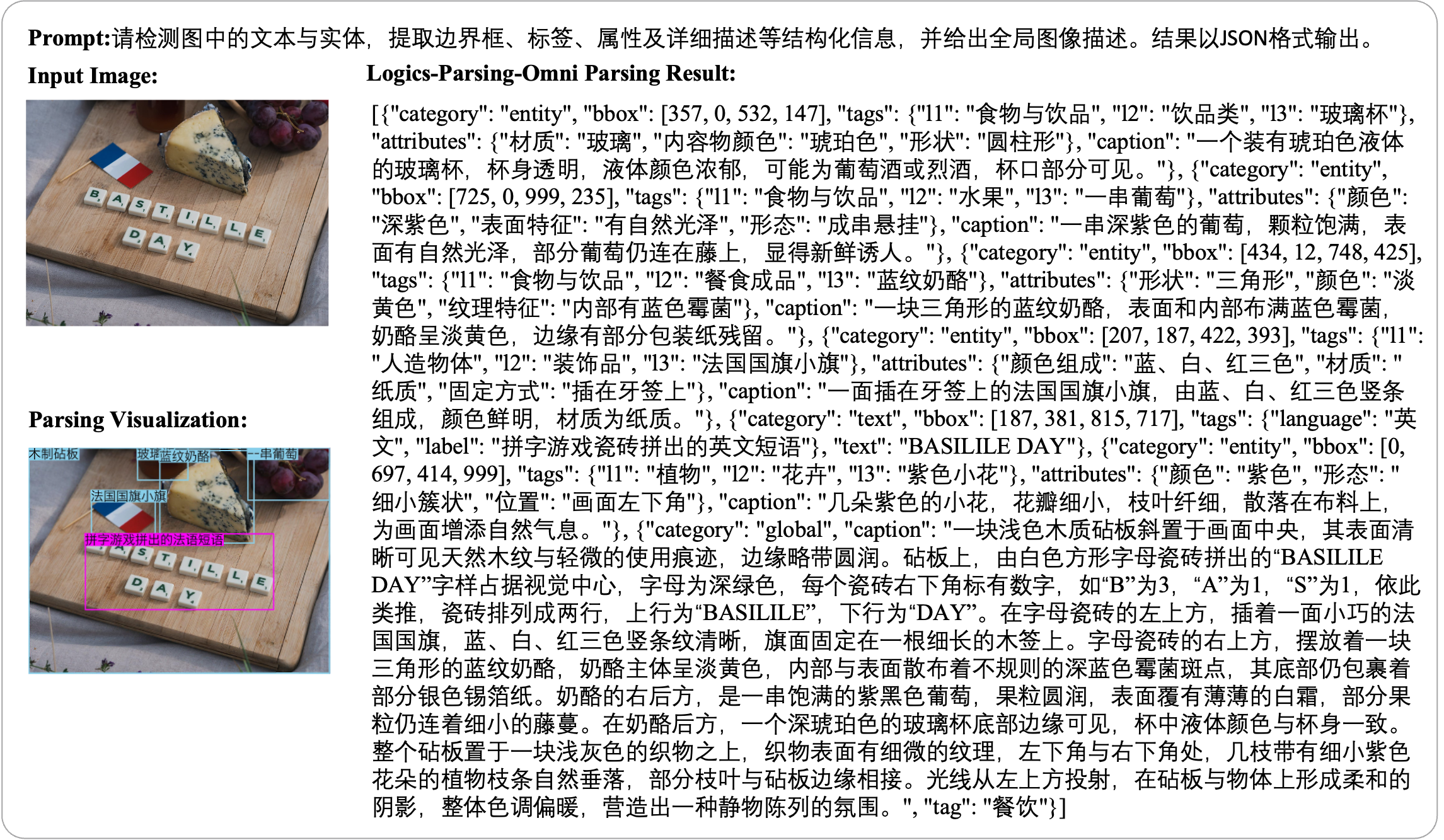}
    \includegraphics[width=1.0\textwidth]{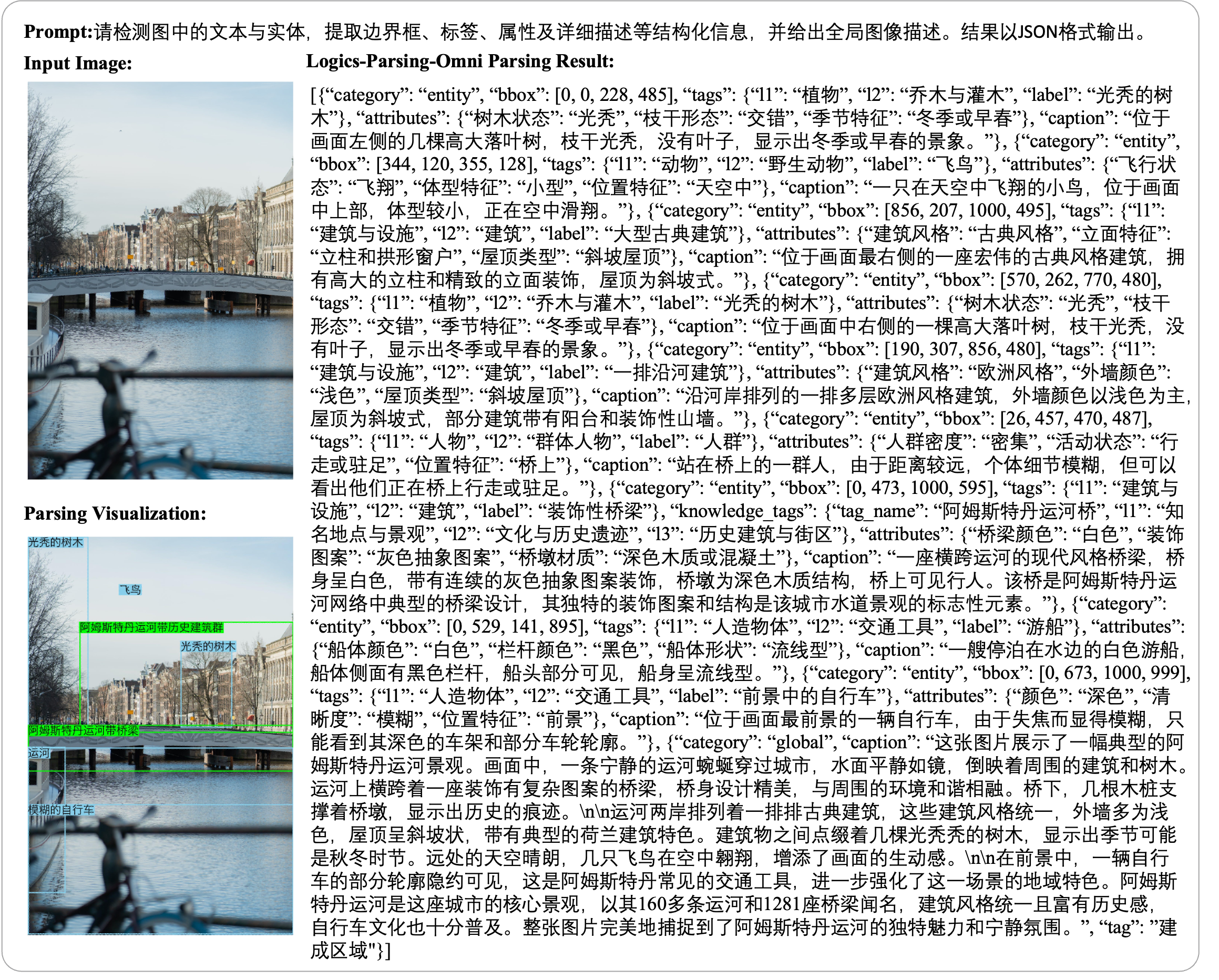}
    \caption{Qualitative examples illustrating the natural image parsing capability of Logics-Parsing-Omni.}
    \label{fig:image_parisng_exps}
\end{figure}

\textit{Benchmark Design.} Fig. \ref{fig:image_parisng_exps} shows 2 image parsing results from Logics-Parsing-Omni sampled from the Natural Image module of OmniParsingBench. Current evaluation faces two inherent constraints: (1) Localization recall for entities and text is sensitive to annotation ambiguities—such as whether closely spaced objects constitute single or multiple entities, or how fragmented/multi-line text should be segmented. These subjective boundaries may induce false negatives in matching-based evaluation, suggesting a need for uncertainty-aware metrics or soft-matching protocols in future benchmarks. (2) The framework emphasizes recall-oriented assessment but lacks explicit quantification of hallucination (i.e., spurious entity/text generation). Integrating precision-focused metrics and false-positive analysis will enable more holistic reliability evaluation.

\textit{Model Scope.} Knowledge-aware parsing remains challenging, as reflected in moderate accuracy for knowledge entity recognition—highlighting the need for more scalable knowledge integration strategies. Furthermore, the current parsing schema prioritizes objective elements (entities, text), while higher-level semantic dimensions—such as aesthetic quality, scene narrative coherence, emotional resonance, and technical image attributes remain under-explored. Future iterations of structured parsing should expand the semantic scope to encompass these dimensions, fostering a more comprehensive understanding of visual content. These directions will guide both benchmark evolution and model development toward holistic image interpretation.

\subsection{Detailed Evaluation on the Graphics Module of OmniParsingBench}
\label{sec:appendix_graphics}
As discussed in the main text, we systematically merged the evaluation metrics into Perception and Cognition dimensions to provide a macroscopic view of the model's graphics capabilities. In this section, we present the fine-grained, deconstructed evaluation results for the Chart and Geometry domains, providing deeper insights into the specific strengths of Logics-Parsing-Omni. To clearly bridge the unified graphics results in the main text (Table~\ref{tab:Graphics_exp}) with the granular metrics presented below, we introduce the exact cross-domain aggregation rules as follows:

\vspace{1mm}
\noindent\textbf{Perception:}
\begin{itemize}[topsep=2pt, itemsep=3pt, parsep=0pt, leftmargin=*]
    \item \textbf{Location:} Represents geometric spatial positioning, derived from the average of coordinate and radius metrics in the Geometry domain. \\
    $\text{Location} = \frac{1}{2} \left( \text{Geo}_{\text{Coord}} + \text{Geo}_{\text{Radius}} \right)$
    
    \item \textbf{Content - Element:} Evaluates the exact existence and recall of fundamental geometric elements. \\
    $\text{Element} = \frac{1}{3} \left( \text{Geo}_{\text{Point}} + \text{Geo}_{\text{Line/Arc}} + \text{Geo}_{\text{Circle}} \right)$
    
    \item \textbf{Content - Relation:} Derived exclusively from explicit structural and quantitative relationships in geometry. \\
    $\text{Relation} = \frac{1}{2} \left(\text{Geo}_{\text{Geo. Rel.}} + \text{Geo}_{\text{Quant. Rel.}}\right)$
    
    \item \textbf{Content - Chart:} Represents the exactness of HTML structural parsing in the Chart domain, directly adopting the structural F1-Score. \\
    $\text{Chart} = \text{Chart}_{\text{Avg.}}$
    
    \item \textbf{Avg.:} The macro-average of the four detailed perception metrics above. \\
    $\text{Avg.} = \frac{1}{4} \left(\text{Location} + \text{Element} + \text{Relation} + \text{Chart}\right)$
\end{itemize}

\vspace{1mm}
\noindent\textbf{Cognition:}
\begin{itemize}[topsep=2pt, itemsep=3pt, parsep=0pt, leftmargin=*]
    \item \textbf{Element:} Represents the fundamental information extraction capability across both domains. \\
    $\text{Element} = \frac{1}{2} \left(\text{Chart}_{\text{OCR \& Data}} + \text{Geo}_{\text{Basic Elements}}\right)$
    
    \item \textbf{Relation:} Captures the ability to extract and bind topological and visual structures. \\
    $\text{Relation} = \frac{1}{3} \left(\text{Chart}_{\text{Visual Elements}} + \text{Geo}_{\text{Geometric Rel.}} + \text{Geo}_{\text{Quantitative Rel.}}\right)$
    
    \item \textbf{Reasoning:} Represents complex logical deduction via implicit chart trends. \\
    $\text{Reasoning} = \text{Chart}_{\text{Logical Reasoning}}$
    
    \item \textbf{Avg.:} The exact mean of these three cognitive dimensions. \\
    $\text{Avg.} = \frac{1}{3} \left(\text{Element} + \text{Relation} + \text{Reasoning}\right)$
\end{itemize}
\vspace{2mm}

\paragraph{Chart.}
Table~\ref{tab:chart_exp} details the models' performance on the Chart sub-benchmark. The Perception phase evaluates the exactness of structural reverse-rendering across diverse visual representations. Specifically, it measures the data extraction accuracy for statistical charts (e.g., bar and line charts) using the F1-score (Stat. Charts), and assesses the topological correctness of structural diagrams (e.g., flowcharts) via sequence similarity (Struct. Diagrams). Concurrently, the Cognition phase assesses the semantic coverage and reasoning capabilities over these extracted structures.

As shown in Table~\ref{tab:chart_exp}, Logics-Parsing-Omni attains an overall score of 91.71, an absolute gain of 7.91 over the baseline (Qwen3-Omni-30B-A3B). This leap is primarily driven by its profound cognitive capabilities, where our model exhibits absolute dominance in fine-grained data comprehension and visual fact grounding. By enforcing strict structural alignment, Logics-Parsing-Omni achieves outstanding performance in both \textit{OCR \& Data} (97.88) and \textit{Visual Elements} (95.81).

\begin{table*}[htbp]
\centering
\caption{Detailed Evaluation Results on the Graphics Module of OmniParsingBench - Chart.}
\label{tab:chart_exp}
\resizebox{\textwidth}{!}{%
\begin{tabular}{l c ccc cccc}
\toprule
\multicolumn{1}{c}{\multirow{2}{*}{\textbf{Model}}} & \multirow{2}{*}{\textbf{Overall}} & \multicolumn{3}{c}{\textbf{Perception}} & \multicolumn{4}{c}{\textbf{Cognition}} \\
\cmidrule(lr){3-5} \cmidrule(lr){6-9}
 & & Stat. Charts (F1) & Struct. Diagrams (Sim.) & Avg. & OCR \& Data & Visual Elements & Logical Reasoning & Avg. Acc. \\
\midrule
Gemini-3-Pro~\footnotemark[13]            & 88.00 & 87.43 & 89.60 & 87.80 & 89.22 & 90.70 & 83.11 & 88.20 \\
GPT-5.2~\footnotemark[9]                  & 88.28 & 79.63 & 87.49 & 83.56 & 94.88 & 87.21 & \textbf{93.95} & 93.00 \\
Qwen3.5-397B-A17B~\footnotemark[16]       & 87.41 & \textbf{92.28} & \textbf{90.12} & \textbf{91.91} & 83.57 & 90.23 & 73.97 & 82.90 \\
Qwen3-VL-235B-A22B~\cite{Qwen3-VL}        & 86.05 & 84.51 & 81.46 & 83.99 & 89.22 & 91.63 & 81.74 & 88.10 \\
Qwen3-VL-30B-A3B~\cite{Qwen3-VL}          & 75.90 & 63.22 & 87.76 & 67.40 & 86.75 & 88.37 & 74.43 & 84.40 \\
Qwen3-Omni-30B-A3B~\cite{Qwen3-Omni}      & 83.80 & 88.08 & 76.97 & 86.19 & 82.33 & 86.51 & 73.97 & 81.40  \\
\midrule
\textbf{Logics-Parsing-Omni (Ours)}       & \textbf{91.71} & 88.13 & 88.06 & 88.12 & \textbf{97.88} & \textbf{95.81} & 88.13 & \textbf{95.30} \\
\bottomrule
\end{tabular}%
}
\end{table*}

\paragraph{Geometry.}
Table~\ref{tab:geo_exp} presents the model's performance on the Geometry sub-benchmark. To rigorously assess the model's capability in decoupling complex spatial topologies and mathematical constraints, the Perception phase within this specific domain employs a specialized weighted metric for its internal \textit{Avg. Score}: exact positioning of coordinates and radii (30\% weight), verifying the existence of basic elements like points, lines/arcs, and circles (30\% weight), and extracting explicit relations encompassing geometric topologies and quantitative annotations (40\% weight). Concurrently, the Cognition evaluation adopts a semantic-coverage QA framework designed to assess the understanding of basic elements, geometric relations (e.g., parallelism, collinearity), and quantitative relations.

For geometry, our model's overall score improved from 71.12 to 85.60, a significant improvement primarily attributed to its fine-grained Perception layer. Logics-Parsing-Omni achieves an exceptional 80.97 in Coordinate Positioning, marking a massive leap from the baseline (32.71) and vastly outperforming Gemini-3-Pro (72.53). Anchored by this precise spatial perception, our model attains the highest average Cognition accuracy (92.00). It excels particularly in the semantic extraction of Basic Elements (95.25) and Quantitative Relations (94.85).

\begin{table*}[h]
    \centering
    \caption{Detailed evaluation results on the Graphics Module of OmniParsingBench - Geometrics.}
    \label{tab:geo_exp}
    \renewcommand{\arraystretch}{1.2} 
    \setlength{\tabcolsep}{4pt} 
    \resizebox{\textwidth}{!}{%
    \begin{tabular}{l c cc ccc cc c cccc}
    \toprule
    \multicolumn{1}{c}{\multirow{4}{*}{\textbf{Model}}} & \multirow{4}{*}{\textbf{Overall}} & \multicolumn{8}{c}{\textbf{Perception}} & \multicolumn{4}{c}{\textbf{Cognition}} \\
    \cmidrule(lr){3-10} \cmidrule(lr){11-14}
    & & \multicolumn{2}{c}{Positioning} & \multicolumn{3}{c}{Existence} & \multicolumn{2}{c}{Relation} & \multirow{2}{*}{\makecell{Avg.}} & \multirow{2}{*}{\makecell{Basic \\ Elements}} & \multirow{2}{*}{\makecell{Geometric \\ Relations}} & \multirow{2}{*}{\makecell{Quantitative \\ Relations}} & \multirow{2}{*}{\makecell{Avg.}} \\
    \cmidrule(lr){3-4} \cmidrule(lr){5-7} \cmidrule(lr){8-9}

    & & Coord. & Radius & Point & Line/Arc & Circle & Geo. & Quant. & & & & & \\
    \midrule
    
    Gemini-3-Pro~\footnotemark[13]       & \textbf{86.05} & 72.53 & 85.20 & \textbf{92.63} & \textbf{69.04} & 96.97 & \textbf{77.88} & \textbf{90.02} & \textbf{83.10} & 89.51 & \textbf{87.03} & 91.75 & 89.00 \\
    GPT-5.2~\footnotemark[9]             & 77.13 & 5.26 & 85.98 & 89.63 & 62.93 & \textbf{97.33} & 58.45 & 75.49 & 65.46 & 91.48 & 81.91 & 92.78 & 88.80 \\
    Qwen3.5-397B-A17B~\footnotemark[16]  & 78.21 & 7.30 & 85.66 & 88.87 & 61.17 & 96.40 & 68.95 & 80.15 & 68.41 & 88.85 & 84.64 & 92.78 & 88.00 \\
    Qwen3-VL-235B-A22B~\cite{Qwen3-VL}   & 72.92 & 16.75 & 88.37 & 89.89 & 62.91 & 96.80 & 50.77 & 81.79 & 67.24 & 82.13 & 67.58 & 89.69 & 78.60 \\
    Qwen3-VL-30B-A3B~\cite{Qwen3-VL}     & 70.60 & 37.90 & 86.15 & 86.87 & 59.24 & 96.57 & 37.85 & 67.22 & 63.89 & 81.15 & 67.24 & 83.51 & 77.30 \\
    Qwen3-Omni-30B-A3B~\cite{Qwen3-Omni} & 71.12 & 32.71 & 87.17 & 88.56 & 59.65 & 97.17 & 41.73 & 68.39 & 64.54 & 80.16 & 71.67 & 80.41 & 77.70 \\
    \midrule
    \textbf{Logics-Parsing-Omni (Ours)}  & 85.60 & \textbf{80.97} & \textbf{89.94} & 91.84 & 60.40 & 94.73 & 61.34 & 82.96 & 79.19 & \textbf{95.25} & 84.30 & \textbf{94.85} & \textbf{92.00} \\
    
    \bottomrule
    \end{tabular}
    }
\end{table*}

\begin{table*}[htbp]
\centering
\begin{threeparttable} 
\caption{Detailed evaluation results on the Natural Videos Module of OmniParsingBench - Camera.}
\label{tab:camera_motion_results}
\renewcommand{\arraystretch}{1.3}
\fontsize{8}{8}\selectfont 
\begin{tabular}{l c ccc c}
\toprule
\multirow{2}{*}{\textbf{Model}} & \multirow{2}{*}{\textbf{Overall}} & \multicolumn{3}{c}{\textbf{Perception}} & \textbf{Cognition} \\
\cmidrule(lr){3-5} \cmidrule(lr){6-6}
& & Joint Acc. & L1 Acc. & L2 Acc. & Acc. \\
\midrule
Gemini-3-Pro        & 38.37 & \underline{34.81} & \underline{65.19} & 34.81 & 41.92 \\
Qwen3-Omni-30B-A3B  & \underline{42.50} & 34.62 & 63.08 & \underline{36.54} & \underline{50.38} \\
Logics-Parsing-Omni & \textbf{67.12} & \textbf{65.58} & \textbf{88.65} & \textbf{65.58} & \textbf{68.65} \\
\bottomrule
\end{tabular}
\begin{tablenotes}
\fontsize{7}{7}\selectfont 
\item[] \textit{Note:} Bold text indicates the best result, and underlined text indicates the second-best result.
\end{tablenotes}
\end{threeparttable}
\end{table*}

\subsection{Detailed Evaluation on the Aduio Module of OmniParsingBench}
\label{sec:appendix_audios}
The exact calculation rules of (Table~\ref{tab:logicsaudiobench}) as follows:

\vspace{1mm}
\noindent\textbf{Perception:}
\begin{itemize}[topsep=2pt, itemsep=3pt, parsep=0pt, leftmargin=*]
    \item \textbf{Avg.:} Aggregates the fundamental audio perception metrics, specifically Time F1, Key F1, and ASR WER. \\
    $\text{Avg.} = \frac{1}{3} \big(\text{Time} + \text{Key} + (1 - \min(1, \text{ASR})) \times 100 \big)$
\end{itemize}

\vspace{1mm}
\noindent\textbf{Cognition:}
\begin{itemize}[topsep=2pt, itemsep=3pt, parsep=0pt, leftmargin=*]
    \item \textbf{Avg.:} Represents the overall cognitive capability, calculated as the exact mean of Information Extraction Accuracy and Recognition Accuracy. \\
    $\text{Avg.} = \frac{1}{2} (\text{Info.\ Ext.} + \text{Rec.})$
\end{itemize}

\vspace{1mm}
\noindent\textbf{Overall:}
\begin{itemize}[topsep=2pt, itemsep=3pt, parsep=0pt, leftmargin=*]
    \item \textbf{Overall:} Calculated as the unweighted mean of the two primary dimensions, reflecting the model's comprehensive performance. \\
    $\text{Overall} = \frac{1}{2} (\text{Perception}_{\text{avg}} + \text{Cognition}_{\text{avg}})$
\end{itemize}

\subsection{Detailed Evaluation on the Natural Video Module of OmniParsingBench}
\label{sec:appendix_videos}
The exact calculation rules of (Table~\ref{tab:logics_natural_videos_combined}) as follows:

\vspace{1mm}
\noindent\textbf{Perception:}
\begin{itemize}[topsep=2pt, itemsep=3pt, parsep=0pt, leftmargin=*]
    \item \textbf{Avg.:} Aggregates all video perception metrics, specifically Time F1, Key F1, ASR WER, Visual accuracy, and Camera accuracy. \\
    $\text{Avg.} = \frac{1}{5} \big(\text{Time} + \text{Key} + (1 - \min(1, \text{ASR})) \times 100 + \text{Visual} + \text{Camera} \big) $
\end{itemize}

\vspace{1mm}
\noindent\textbf{Cognition:}
\begin{itemize}[topsep=2pt, itemsep=3pt, parsep=0pt, leftmargin=*]
    \item \textbf{Avg.:} Represents the mean accuracy across all cognition sub-categories \\
    $\text{Avg.} = \frac{1}{8} (\text{Count} + \text{Rec.} + \text{Info. Ext.} + \text{Attr.\&States} + \text{Event} + \text{Camera} + \text{Emot.} + \text{Caus. Rsn.})$
\end{itemize}

\vspace{1mm}
\noindent\textbf{Overall:}
\begin{itemize}[topsep=2pt, itemsep=3pt, parsep=0pt, leftmargin=*]
    \item \textbf{Overall:} Calculated as the unweighted mean of the two primary dimensions, reflecting the model's comprehensive performance. \\
    $\text{Overall} = \frac{1}{2} (\text{Perception}_{\text{avg}} + \text{Cognition}_{\text{avg}})$
\end{itemize}

Table~\ref{tab:camera_motion_results} details the models’ performance on the Camera sub-benchmark. In perception, it attains a joint accuracy of 65.58\%, significantly outperforming Gemini-3-Pro and Qwen3-Omni-30B-A3B, especially in fine-grained L2 classification. In semantic cognition, the model reaches 68.65\% accuracy (+18.27\% over the strongest baseline), demonstrating superior integration of camera language into multimodal narratives. Overall, it reaches a leading score of 67.12\% (calculated as the average of perception joint accuracy and cognition accuracy).

\subsection{Detailed Evaluation on the Text-Rich Video Module of OmniParsingBench}
\label{sec:appendix_text_rich_videos}
The exact calculation rules of (Table~\ref{tab:course_parsing_results}) as follows:

\vspace{1mm}
\noindent\textbf{Perception:}
\begin{itemize}[topsep=2pt, itemsep=3pt, parsep=0pt, leftmargin=*]
    \item \textbf{Avg.:} Aggregates all video perception metrics, specifically Time F1, ASR WER, and Visual OCR Score. \\
    $\text{Avg.} = \frac{1}{3} \big(\text{Time} + \text{Visual} + (1 - \min(1, \text{ASR})) \times 100  \big) $
\end{itemize}

\vspace{1mm}
\noindent\textbf{Cognition:}
\begin{itemize}[topsep=2pt, itemsep=3pt, parsep=0pt, leftmargin=*]
    \item \textbf{Avg.:} Represents the mean between the In-Depth Str. Cap. score and the average accuracy of all other Cognition categories. \\
    $\text{Avg.} = \frac{1}{2} \big(\text{In-Depth Str. Cap.} + \big(\frac{1}{7} (\text{Info. Ext.} + \text{Object} + \text{OCR} + \text{Attr.\&States} + \text{Event} + \text{Info. Syn.} + \text{Caus. Rsn.}) \big) \big)$
\end{itemize}

\vspace{1mm}
\noindent\textbf{Overall:}
\begin{itemize}[topsep=2pt, itemsep=3pt, parsep=0pt, leftmargin=*]
    \item \textbf{Overall:} Calculated as the unweighted mean of the two primary dimensions, reflecting the model's comprehensive performance. \\
    $\text{Overall} = \frac{1}{2} (\text{Perception}_{\text{avg}} + \text{Cognition}_{\text{avg}})$
\end{itemize}

\subsection{Qualitative Demonstrations of Omni-Modal Parsing}
In this subsection, we present a comprehensive series of qualitative examples to demonstrate the multifaceted capabilities of Logics-Parsing-Omni. These showcases span our proposed unified taxonomy, encompassing diverse and knowledge-intensive scenarios: from natural image understanding and graphics parsing (e.g., complex charts and geometric illustrations), to multi-image difference reasoning, dynamic audio-visual stream parsing (including natural and text-rich educational videos), and intricate document structural analysis.

\begin{figure}[htbp]
    \centering
    \includegraphics[width=1.0\textwidth]{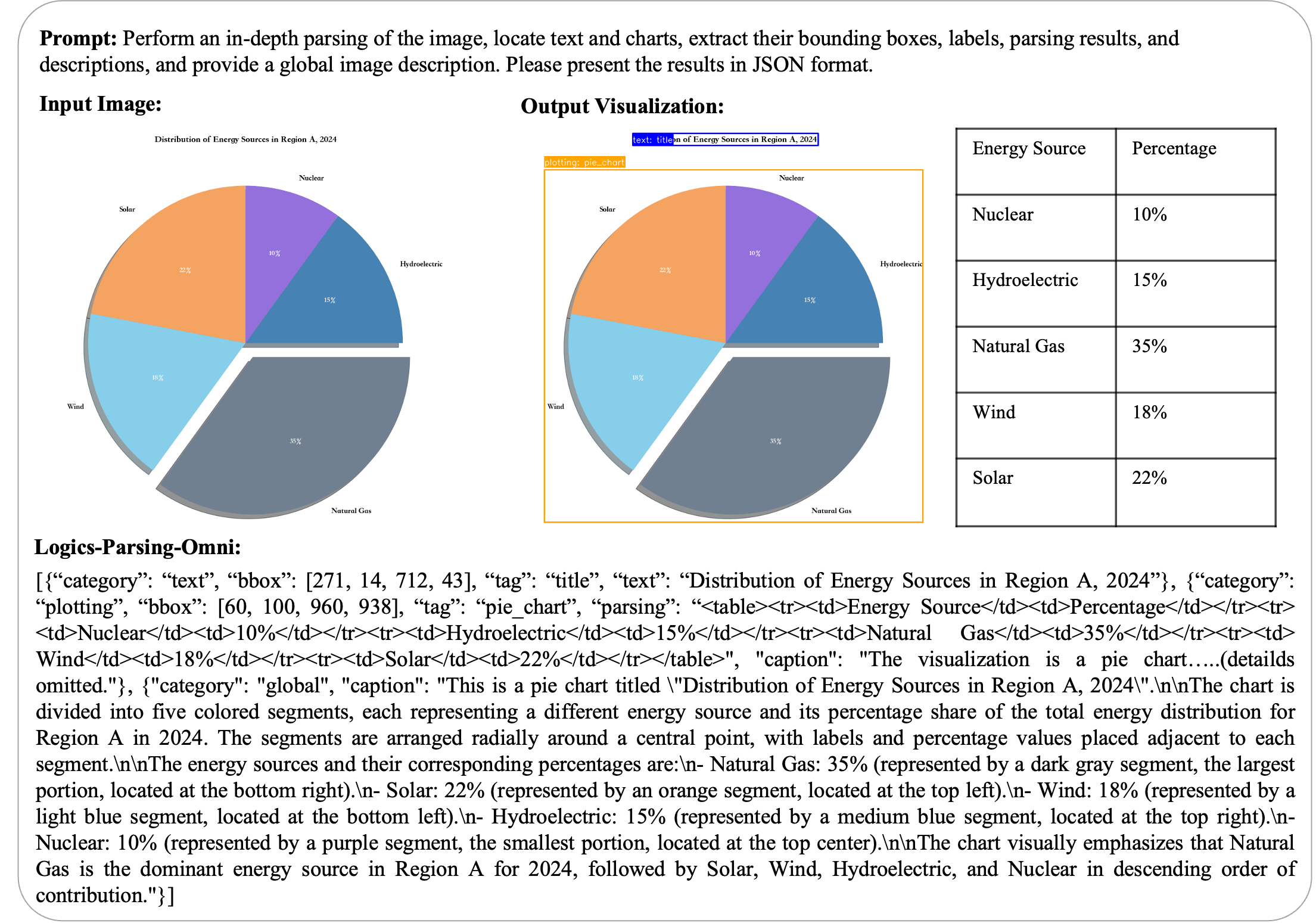}
    \includegraphics[width=1.0\textwidth]{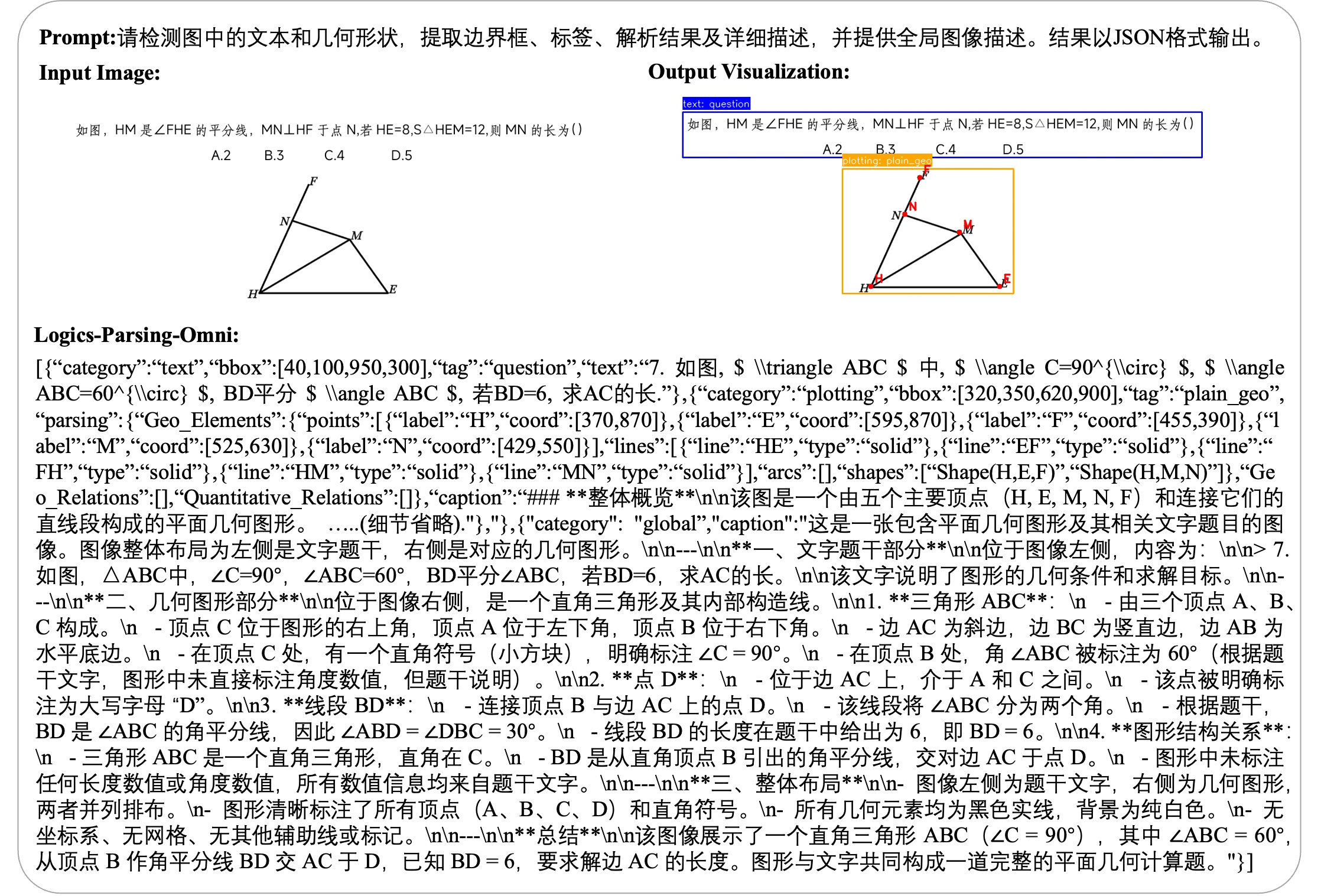}
    \caption{Qualitative examples illustrating the graphics image parsing capability of Logics-Parsing-Omni.}
    \label{fig:image_graphics_cases}
\end{figure}

\begin{figure}[htbp]
    \centering
    \includegraphics[width=1.0\textwidth]{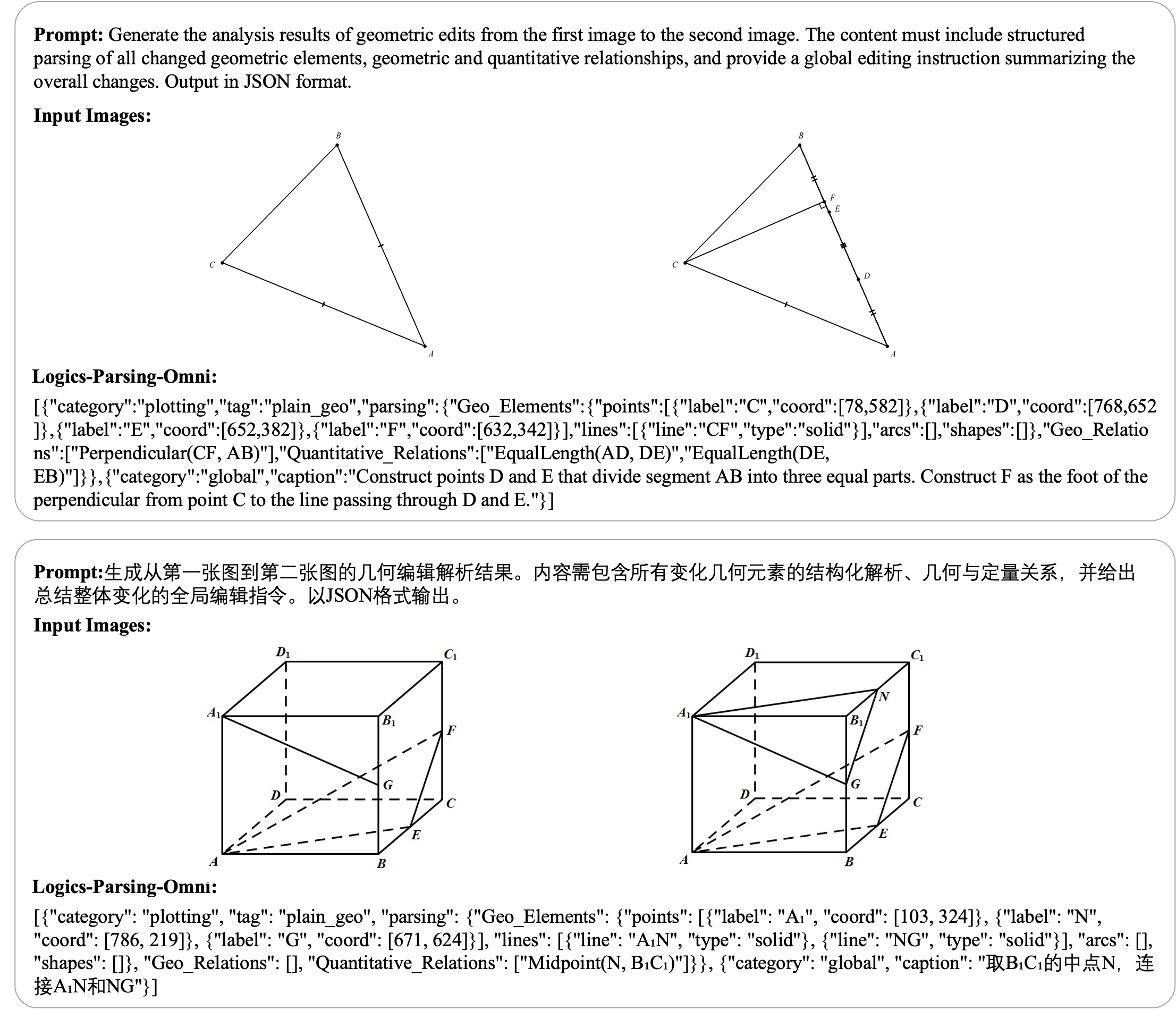}
    \caption{Qualitative examples illustrating the graphics image difference parsing capability of Logics-Parsing-Omni.}
    \label{fig:case1}
\end{figure}

\begin{figure}[htbp]
    \centering
    \includegraphics[width=1.0\textwidth]{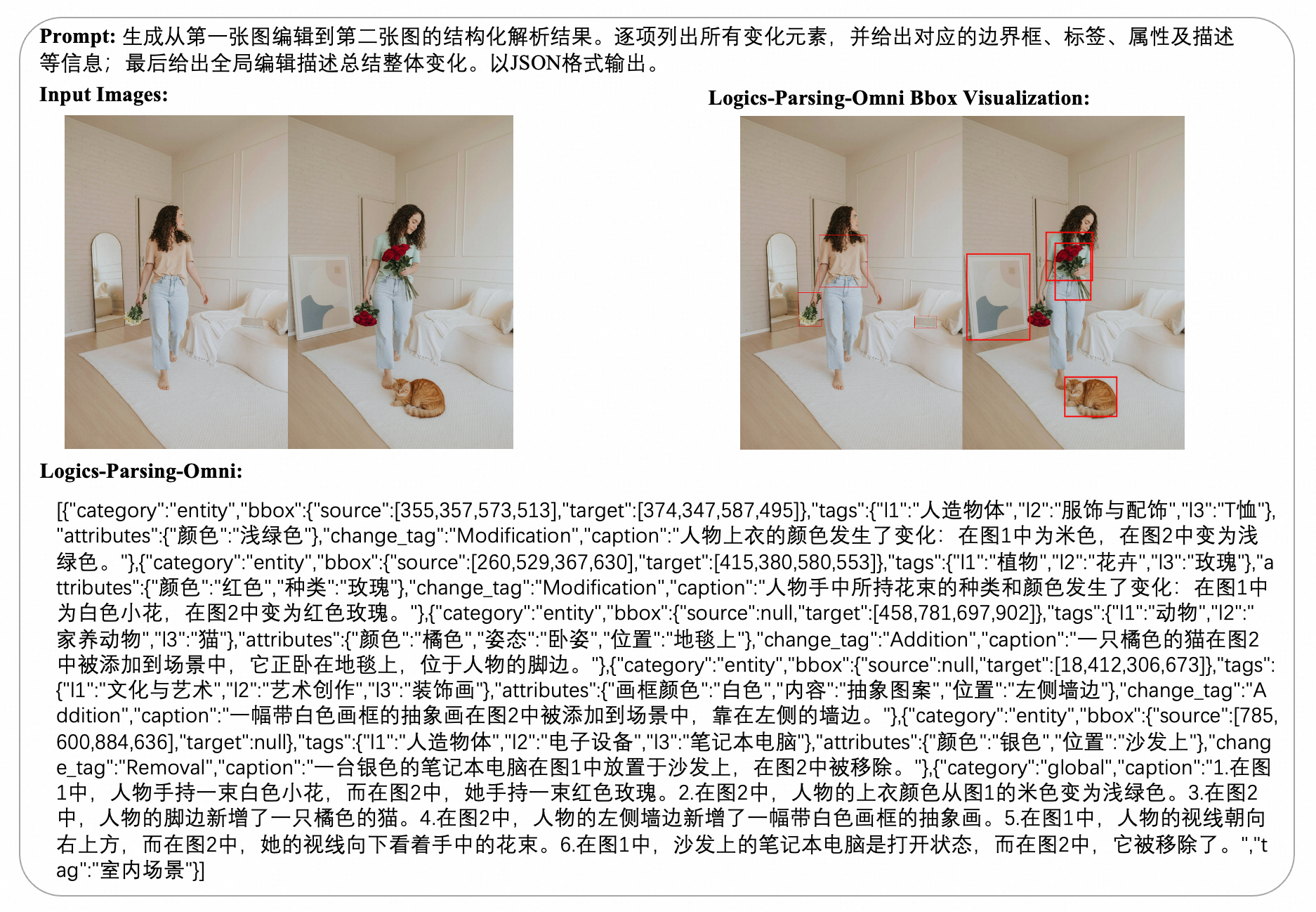}
    \caption{Qualitative examples illustrating the natural image difference parsing capability of Logics-Parsing-Omni.}
    \label{fig:diff_parsing_view_2}
\end{figure}

\begin{figure}[htbp]
    \centering
    \includegraphics[width=1.0\textwidth]{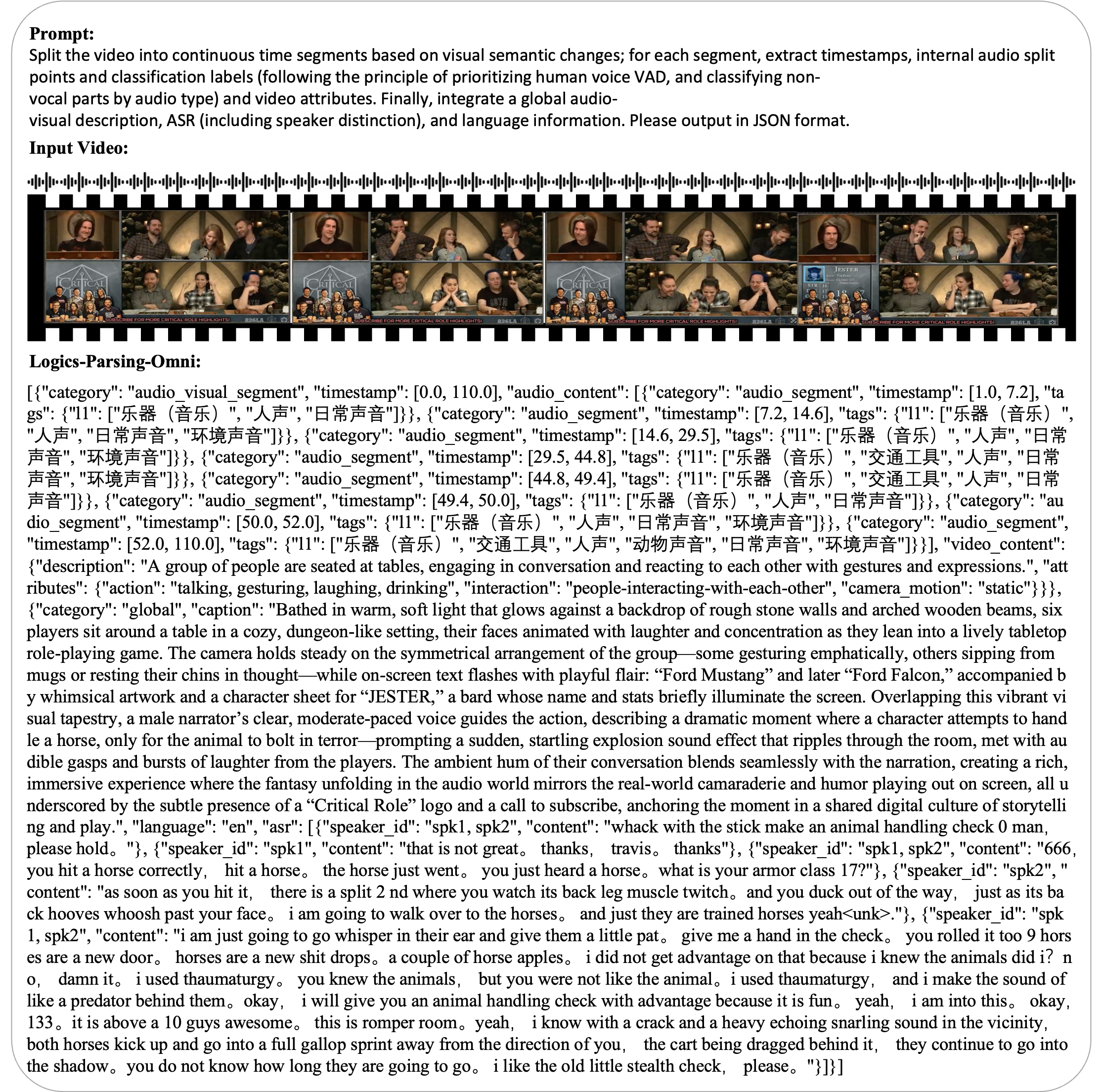}
    \caption{Qualitative examples illustrating the natural video parsing capability of Logics-Parsing-Omni.}
    \label{fig:video_showcase}
\end{figure}

\begin{figure}[htbp]
    \centering
    \includegraphics[width=1.0\textwidth]{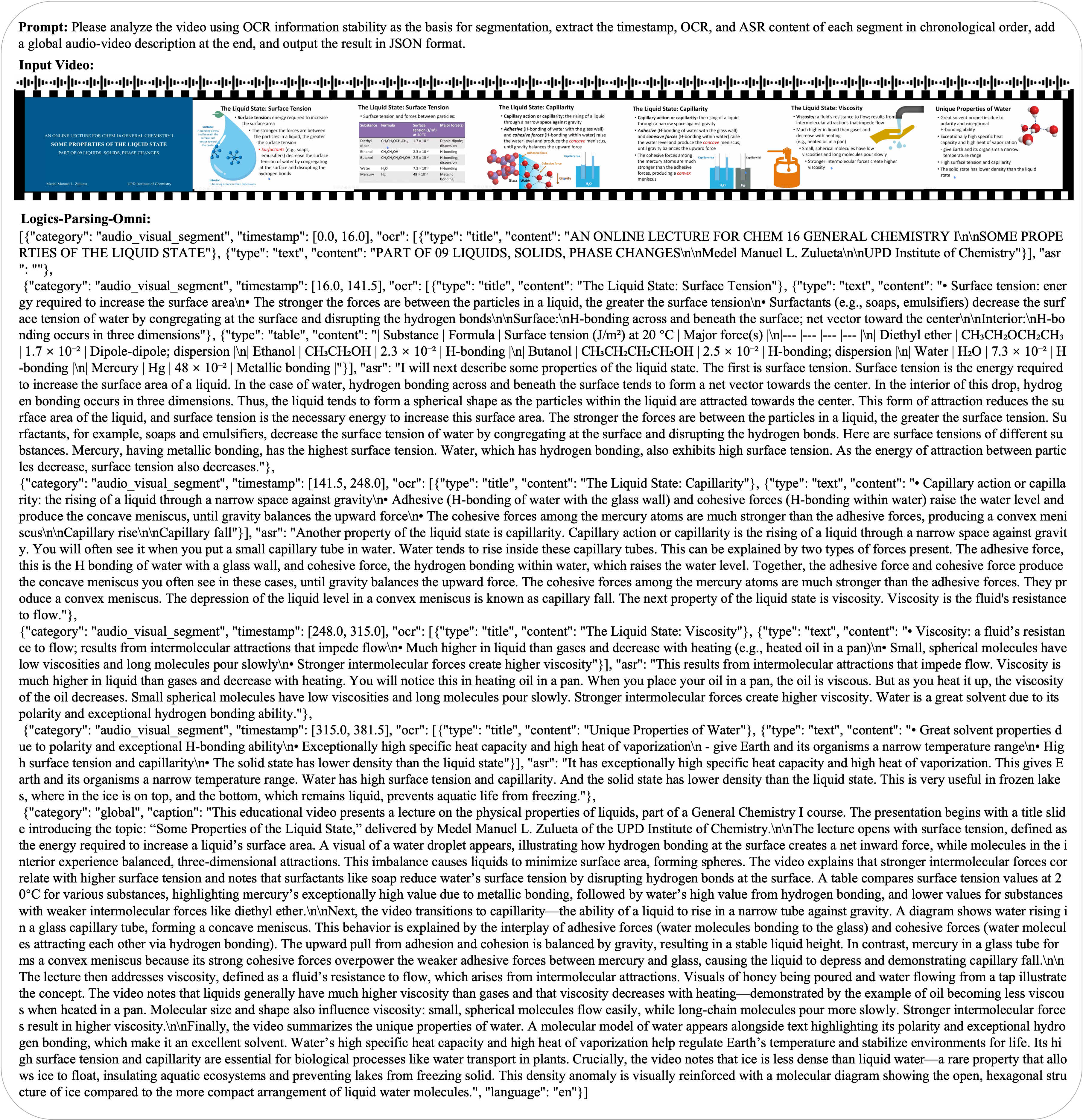}
    \caption{Qualitative example illustrating the comprehensive text-rich video parsing capability of Logics-Parsing-Omni.}
    \label{fig:course_video_parsing_showcase}
    
\end{figure}

\begin{figure}[htbp]
    \centering
    \includegraphics[width=1.0\textwidth]{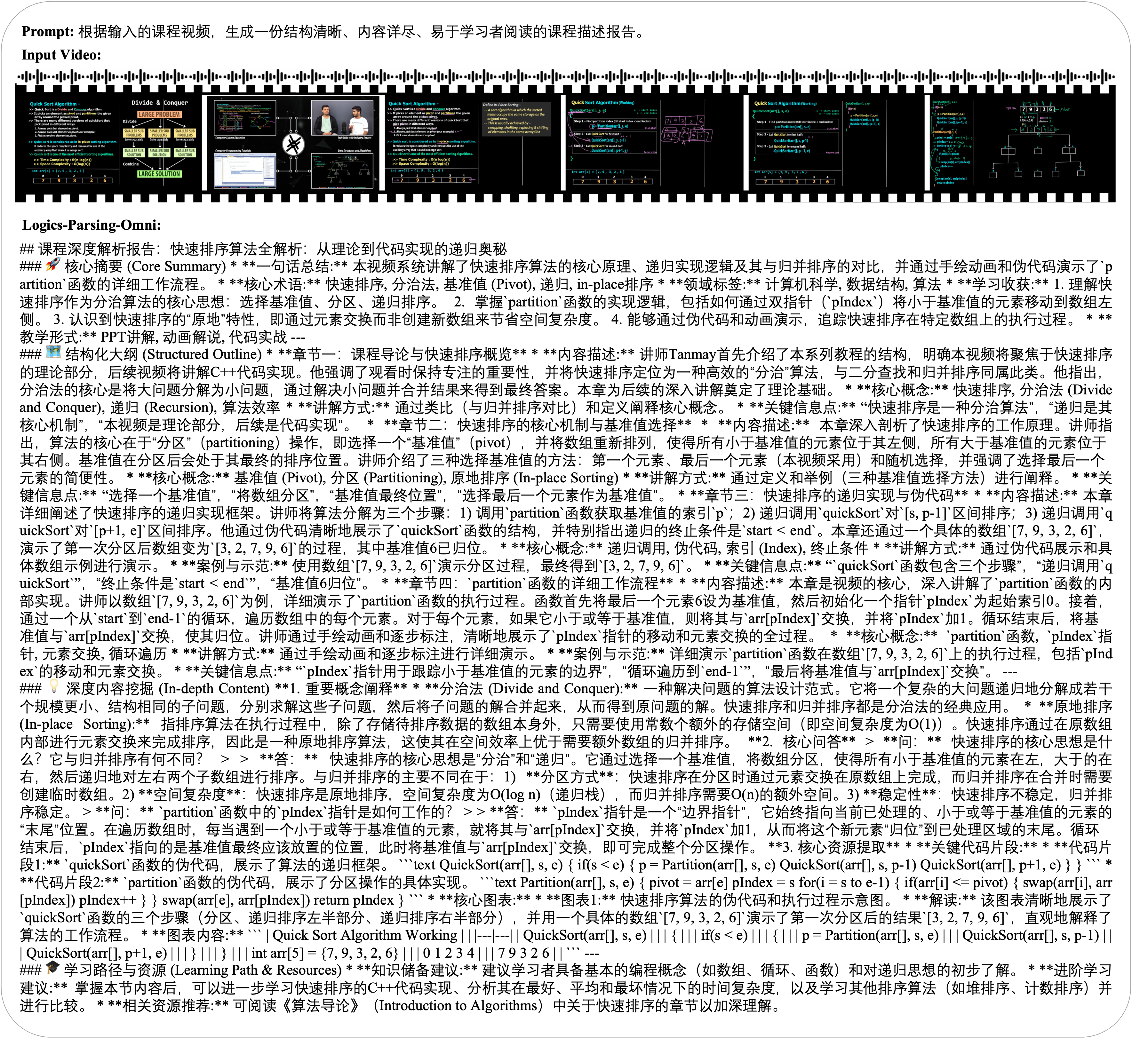}
    \caption{Qualitative examples illustrating the text-rich video in-depth caption capability of Logics-Parsing-Omni.}
    \label{fig:course_video_indepth_caption_showcase}
\end{figure}

\begin{figure}[htbp]
    \centering
    \includegraphics[width=1.0\textwidth]{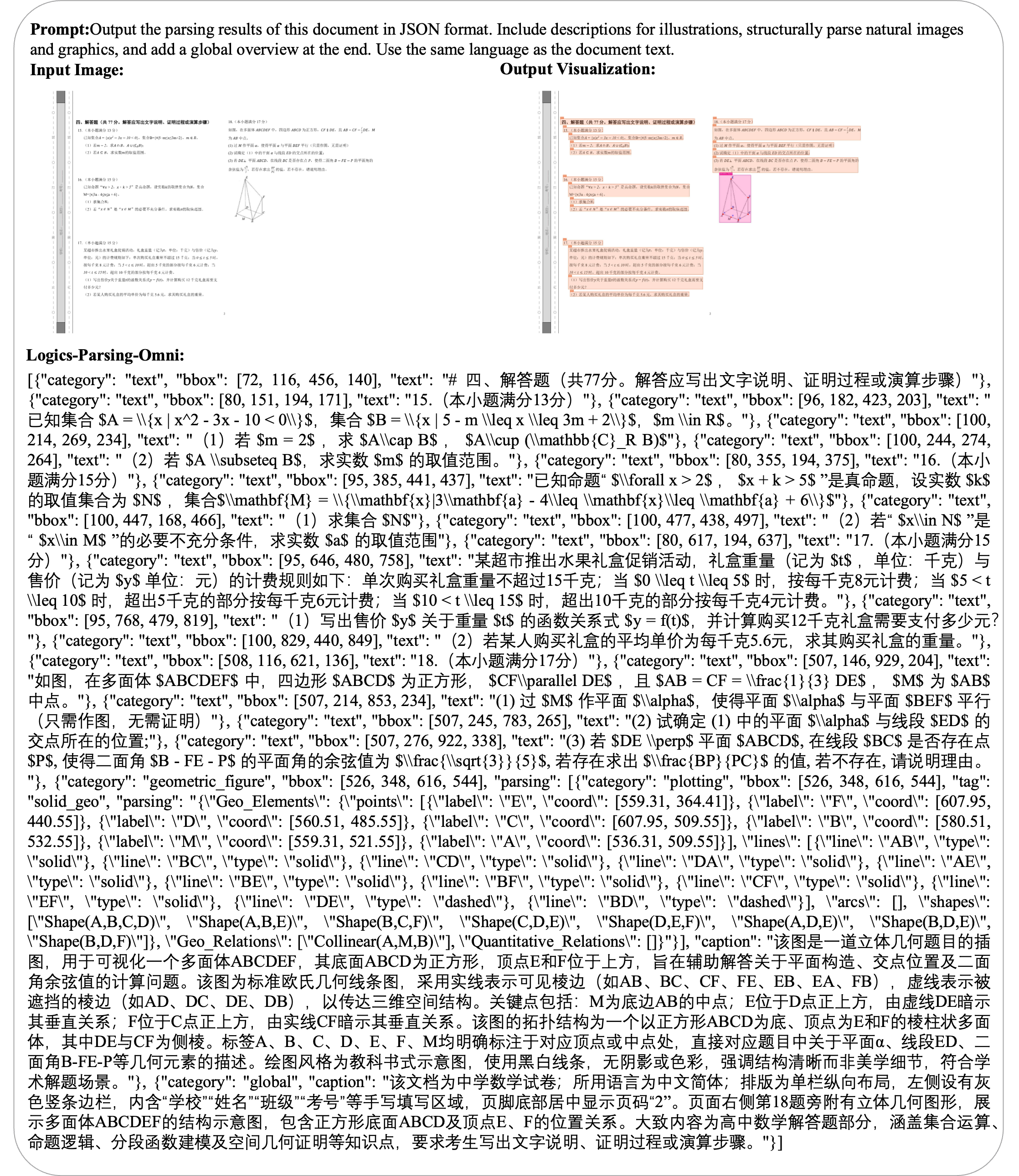}
    \caption{Qualitative example illustrating the comprehensive document structure and semantic parsing capability of Logics-Parsing-Omni on the geometric illustration.}
    \label{fig:img_doc_geo_case}
\end{figure}

\begin{figure}[htbp]
    \centering
    \includegraphics[width=1.0\textwidth]{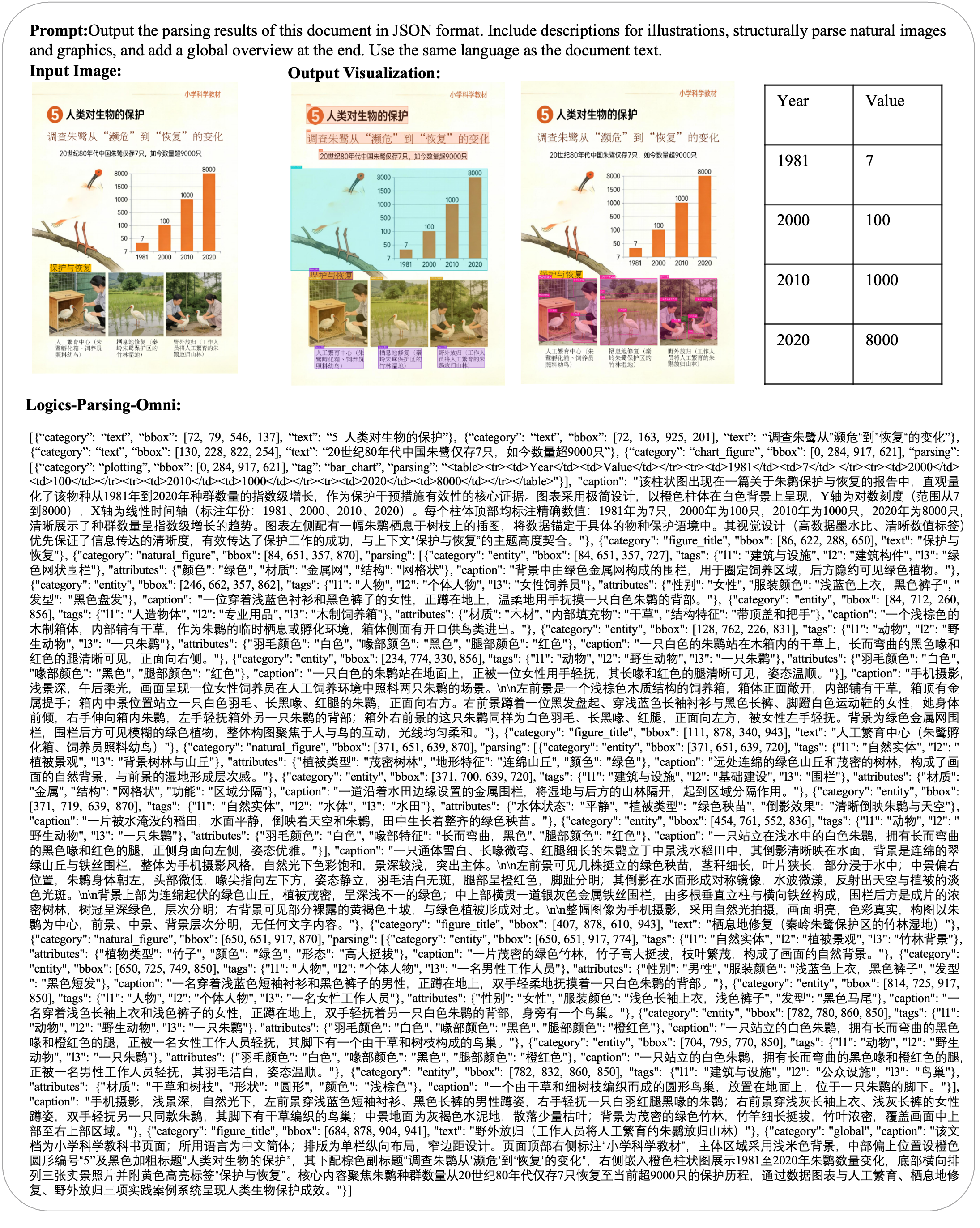}
    
    \caption{Qualitative example illustrating the comprehensive document structure and semantic parsing capability of Logics-Parsing-Omni on the graphics and natrual illustrations.}
    \label{fig:img_doc_chart_case}
\end{figure}

\end{document}